\documentclass[preprint,3p,10pt]{elsarticle}

\usepackage{amssymb}
\usepackage{amsmath}
\usepackage{enumitem}

\usepackage{graphicx}
\usepackage{booktabs}
\usepackage{tcolorbox} 
\usepackage[dvipsnames]{xcolor}
\usepackage{soul}
\usepackage{todonotes}
\usepackage{float}
\usepackage{setspace}
\usepackage{siunitx}
\usepackage{subcaption}
\usepackage[ruled,vlined]{algorithm2e}
\usepackage{nomencl}
\usepackage{tabularx}
\usepackage{makecell}
\usepackage{multirow}
\makenomenclature
\usepackage{etoolbox}
\renewcommand\nomgroup[1]{%
  \item[\bfseries
  \ifstrequal{#1}{S}{List of Symbols}{%
  \ifstrequal{#1}{A}{List of Abbreviations}{%
  \ifstrequal{#1}{O}{Other symbols}{}}}%
]}

\usepackage{hyperref}

\begin{document}

\begin{frontmatter}

\author[delft]{James Josep Perry \fnref{equal}}
\author[delft]{Pablo Garcia-Conde Ortiz \fnref{equal}}
\author[delft]{George Konstantinou \fnref{equal}}
\author[delft]{Cornelie Vergouwen }
\author[delft]{Edlyn Santha Kumaran }

\author[delft]{Morteza Moradi\corref{cor1}}
\ead{m.moradi-1@tudelft.nl}

\cortext[cor1]{Corresponding author}
\fntext[equal]{These authors contributed equally.}

\title{Semi-supervised and unsupervised learning for health indicator extraction from guided waves in aerospace composite structures}

\affiliation[delft]{organization={Center of Excellence in Artificial Intelligence for Structures, Prognostics \& Health Management, Aerospace Engineering Faculty, Delft University of Technology},
            addressline={Kluyverweg 1},
            city={Delft},
            postcode={2629 HS},
            country={the Netherlands}}

\begin{abstract}

\noindent Health indicators (HIs) are central to diagnosing and prognosing the condition of aerospace composite structures, enabling efficient maintenance and operational safety. However, extracting reliable HIs remains challenging due to variability in material properties, stochastic damage evolution, and diverse damage modes. Manufacturing defects (e.g.\ disbonds) and in‑service incidents (e.g.\ bird strikes) further complicate this process. This study presents a comprehensive data‑driven framework that learns HIs via two learning approaches integrated with multi‑domain signal processing. Because ground‑truth HIs are unavailable, a semi-supervised and an unsupervised approach are proposed: (i) a diversity deep semi-supervised anomaly detection (Diversity-DeepSAD) approach augmented with continuous auxiliary labels used as hypothetical damage proxies, which overcomes the limitation of prior binary labels \textcolor{black}{and enables modelling of intermediate degradation,} and (ii) a degradation-trend-constrained variational autoencoder (DTC-VAE), in which the monotonicity criterion is embedded via an explicit trend constraint. Guided waves with multiple excitation frequencies are used to monitor single‑stiffener composite structures under fatigue loading. Time, frequency, and time–frequency representations are explored, and per‑frequency HIs are fused via unsupervised ensemble learning to mitigate frequency dependence and reduce variance. Using fast Fourier transform features, the \textcolor{black}{models achieved fitness scores of 81.6\% (Diversity-DeepSAD) and 92.3\% (DTC-VAE), indicating improved 
monotonicity and consistency over existing baselines.} 
 \textcolor{black}{The proposed history-independent framework, supported by prognostic metrics–guided Bayesian optimisation and excitation frequency-agnostic HI fusion, enables the estimation of more robust HIs for aeronautical composite structures.}

\end{abstract}

\begin{keyword}

Prognostics and health management \sep Health indicator \sep Constrained variational autoencoder \sep Semi-supervised learning \sep Aeronautical composite structures

\end{keyword}

\end{frontmatter}

\begin{spacing}{0.4}

\newpage

\end{spacing}

\begin{tcolorbox}[colframe=black, colback=white, boxrule=1pt, width=\textwidth]
\begin{minipage}[t]{0.47\textwidth}
    \textbf{List of Symbols} \\ \\
    \begin{tabularx}{\textwidth}{@{}lX@{}}
        $c$ & Hypersphere centre \\
        
        $F_{all}$ & Fitness function \\
        $F_{test}$ & Test fitness function \\
        $f$ & Frequency \\
        $H$ & Number of \textcolor{black}{eigenvalues} \\
        $h$ & \textcolor{black}{Eigenvalue number}  \\
        $i, j$ & Timestep \\
        $k$ & Prognostic criteria weighting \\
        $\mathcal{L}$ & Loss function \\
        $M$ & Total number of \textcolor{black}{specimens} \\
        $Mo$ & Monotonicity \\
        $m$ & \textcolor{black}{specimen} number \\
        $N$ & Number of measurements (timesteps) \\
        $N_f$ & Number of frequencies \\
        $n$ & Number of labeled samples \\
        $Pr$ & Prognosability \\
        $Q_{\phi}$ & Encoder probability distribution \\
        $r$ & Rate of degradation \\
        $S$ & Frequency domain statistical feature \\
        $s(f)$ & Frequency domain signal \\
        $t_i$ & Time at timestep $i$ \\
        $Tr$ & Trendability \\
        $u$ & Number of unlabeled samples \\

        $X$ & Time domain statistical feature \\

    \end{tabularx}
\end{minipage}
\hfill
\begin{minipage}[t]{0.47\textwidth}
    \textbf{} \\ \\ 
    \begin{tabularx}{\textwidth}{@{}lX@{}}
    
        $x(i)$ & Time domain signal \\
        $x_i^{m,f}$ & Features at timestep $i$ of specimen $m$ with measurement frequency $f$ \\
        $\hat{x}$ & Reconstructed features \\
        $y_i^{m,f}$ & Health indicator at timestep $i$ of specimen $m$ with measurement frequency $f$\\
        $\tilde{y}$ & \textcolor{black}{Auxiliary} label \\
        $\bar{y}$ & Mean health indicator across folds \\
        $z$ & \textcolor{black}{Autoencoder} latent variables\\
        & \\
        $\alpha$ & Kullback-Leibler divergence weighting\\
        $\beta$ & Reconstruction loss weighting\\
        $\gamma$ & Monotonicity constraint loss weighting\\
        \textcolor{black}{$\epsilon$} & \textcolor{black}{Small number to prevent zero errors}\\
        \textcolor{black}{$\varepsilon$} & \textcolor{black}{Gaussian noise}\\
        \textcolor{black}{$\zeta_h$} & \textcolor{black}{$h^{th}$ eigenvalue of Gram matrix} \\
        $\eta$ & Labelled parameter loss weighting \\
        $\theta$ & Neural network function \\
        $\lambda$ & Diversity loss weighting \\
        $\mu$ & Mean \\
        $\nu$ & L2 regularisation weighting \\
        $\sigma$ & Standard deviation \\
        $\phi$ & Encoder parameters \\
        $\psi$ & Decoder parameters \\
        $\omega_f$ & Weight for frequency $f$ \\

    \end{tabularx}
\end{minipage}
\end{tcolorbox}

\section{Introduction}\label{sec:intro}
\noindent The health of aerospace structures must be continuously monitored to ensure their safety and reliability. An effective monitoring and diagnostic system should detect damage early enough to issue warnings before it escalates into a dangerous situation \cite{jimenez2020towards}. Even earlier diagnostics using damage trend prediction, known as prognostics, can save significant time and cost by enabling timely repairs before a structure reaches an irreparable state \cite{sikorska2011prognostic, badihi2022comprehensive}, in addition to improving reliability \cite{jia2021sample, Guo2024, wu2025multi}. In this context, developing a health indicator (HI) is essential for both diagnostics and prognostics \cite{qiu2003robust, guo2017recurrent, khan2023review, Wei2025}. An HI, being a value which quantifies the health of a structure, serves as a critical bridge between these two aspects towards condition-based maintenance (CBM) \cite{zhang2022marine, huang2024prognostics}. However, deriving a truly comprehensive and accurate HI is exceptionally challenging for most objects, especially engineering systems, due to their inherent complexity \cite{moradi2024designing}.
The comprehensiveness of an HI refers to its ability to account for all potential types of damage that could affect a system's health \cite{moradi2024designing, beaumont2020structural, Moradi2024}. Tracking and quantifying all forms of degradation, even in simple objects made of isotropic materials, is often infeasible. Many degradation processes occur internally, beyond the reach of direct measurement. Although sensors may monitor the effects of different types of damage on the sensed signals, meticulously identifying the damage remains challenging due to numerous known and unknown factors, which depend on the object under monitoring, type of sensors, and structural health monitoring (SHM) techniques employed. This complexity is further exacerbated in the case of advanced systems and materials \cite{beaumont2020structural, ferreira2022remaining}. In summary, claiming to have completely true HI values for any object, especially a complex one like an engineering system, is an overstatement \cite{moradi2024designing, Huang2024}.

While a perfectly accurate HI is impractical, certain principles about HIs hold universally true. For example, the end of life (EoL) represents the complete loss of health in any system. Similarly, a system’s health inevitably decreases over time unless maintenance or self-healing mechanisms are applied. These principles form the foundation for evaluating an HI using key metrics. Prognostability ($Pr$), for instance, ensures that the failure state at EoL is represented by the same HI value across different systems \cite{coble2009identifying, coble2010merging}. Monotonicity ($Mo$) emphasises the consistency of health degradation over time \cite{saxena2008metrics, coble2009identifying, Gonzalez-Muniz2022, Li2023}. \textcolor{black}{Another desirable metric is trendability ($Tr$), which assesses whether the degradation trends of similar systems are consistent. Although uncertainties in manufacturing, loading conditions, and environmental factors may introduce variability, the general trends for similar systems should not diverge significantly. Therefore, ideal HIs for a group of similar systems should satisfy $Mo$, $Pr$, and $Tr$ \cite{coble2010merging, moradi2024designing, Moradi2023Intelligent, moradi2025novel}, all of which are considered in the fitness score.}

As already mentioned, developing or identifying HIs is a challenging task due to the inherent complexity and uncertainty of material degradation processes. This difficulty is amplified in anisotropic materials like composites \cite{senthilkumar2021nondestructive} which are widely used in aerospace applications thanks to their high strength-to-weight ratio and ability to withstand directional stresses \cite{hassani2021structural}. Unlike isotropic materials, composite materials exhibit more complex failure modes arising from their anisotropic properties and the potential interactions between different types of damage \cite{deluca2018, Saeedifar2020, Moradi2024}. These unique characteristics make designing effective HIs for composite structures particularly intricate \cite{beaumont2020structural, Moradi2023Intelligent}. Additionally, uncertainties arising during manufacturing (e.g.\ disbond defects) or operation (e.g.\ bird strikes) exacerbate this complexity \cite{Dienel2019, Aujoux2023}.
To enable in situ monitoring and account for these uncertainties, SHM techniques are essential. An effective SHM technique must be capable of covering most structural areas, particularly critical regions. Among these techniques, guided wave (GW)-based SHM stands out as a practical approach, where multiple sensors are strategically attached to critical regions to monitor the structure’s health. GWs are ultrasonic waves bounded by a structure's material \cite{Aujoux2023} that produce data from which the location and severity of damage to a structure can be determined \cite{deluca2018, Flynn2011, Memmolo2018, Kralovec2020}. This enables SHM in otherwise inaccessible regions of complex composite structures \cite{Janardhan2014}. As GW and most other SHM techniques rely on a network of sensors, model-based approaches face limitations in processing the large volume of complex, high-dimensional sensed signals. Furthermore, noise, multimodality, and the influence of environmental factors make it more challenging to generate HIs from raw GW data \cite{Moradi2024, Yang2020, Yang2023}. Consequently, data-driven and artificial intelligence (AI) approaches offer greater potential for handling such intricate datasets. Due to the unknown true health state of any engineering system, supervised models face limitations in constructing HIs. To address this challenge, semi-supervised or unsupervised AI models can be utilised for data fusion, enabling the generation of effective HIs \cite{Moradi2023SemiSupverised, Li2024}.

Furthermore, the calculation of the typical fitness score, which includes $Mo$, $Pr$, and $Tr$, is prone to bias toward the training units. This bias can lead to potential issues, as the model may produce highly correlated HIs during training, resulting in misleadingly high fitness scores. However, when faced with an unmatched HI from a specific unit during testing, the performance may fall short. To address this, the rectified evaluation criteria \cite{moradi2024designing, Moradi2024} are adopted in this study, ensuring a more reliable assessment during the test phase. This adjustment provides a stronger foundation for comparison and enhances the practical reliability of the standard.

To construct a history-independent HI for aeronautical composite structures, GW-SHM emerges as a promising technique as mentioned earlier. However, each GW measurement requires data collection through a network of PZT sensors, resulting in an enormous volume of data to process. Feeding this data directly into deep learning models demands a high number of input layer nodes, which increases model complexity and computational requirements while reducing interpretability \cite{Yang2023}. A more efficient alternative is to employ signal processing (SP) algorithms as a preprocessing step. SP algorithms often follow explicit solutions, producing globally applicable and faster outcomes. By extracting a set of statistical features from the processed signals, the size and complexity of the models can be significantly reduced. However, identifying which SP method yields the most effective features remains an open question. The proposed frameworks address this by first applying various SP algorithms to extract and compare features in terms of fitness criteria. The selected features are then fed into AI models to generate HIs, ensuring both efficiency and effectiveness in the process. 

\newpage
This work contributes to the field in the following ways:
\textcolor{black}{\begin{enumerate}[label=(\roman*)]
    \item Extension of the Diversity-DeepSAD model into augmented Diversity-DeepSAD by embedding continuous degradation-progression labels instead of binary $\pm1$ labels in its loss function, enabling modelling of intermediate fatigue states essential for composites.
    \item Development of a degradation-trend-constrained variational autoencoder (DTC-VAE) architecture, in which a monotonicity constraint is embedded directly in the latent space to enforce consistent HI evolution, for guided wave monitoring of aerospace composites.
    \item Use of Bayesian hyperparameter optimisation guided by a composite fitness score based on monotonicity, prognosability, and trendability, requiring construction of full run-to-failure HI trajectories for all training and validation units.
    \item Implementation of a frequency-agnostic unsupervised HI fusion mechanism, weighted by composite fitness score, providing a principled alternative to selecting a single excitation frequency in guided wave SHM.
    \item Comprehensive comparison of semi- and unsupervised learning models for generating history-independent HIs, highlighting their behaviour, limitations, and applicability across guided wave SHM scenarios.
\end{enumerate}}

\textcolor{black}{A review of fundamental literature follows in \autoref{sec:lit-study}.} The methodology employed in this study is described in detail in \autoref{sec:method}, with the corresponding results presented in \autoref{sec:results}. A discussion of these findings is provided in \autoref{sec:disc}, culminating in the conclusions outlined in \autoref{sec:conc}.

\section{Literature study} \label{sec:lit-study}
\noindent
\textcolor{black}{
In this section, we first review semi-supervised models relevant to HI generation, followed by a discussion of unsupervised approaches. Semi-supervised models begin with concepts developed from supervised methods. For example, the support vector machine (SVM), introduced by Cortes and Vapnik \cite{cortes1995}, is a fundamental supervised classification algorithm. However, its binary classification ability falls short of generating meaningful HIs, as it ignores the complex gradient between healthy and failure states. Progress toward HI generation came with support vector data description (SVDD) \cite{tax2004} and its deep learning extension, Deep SVDD, proposed by Kim et al.\ in 2015 \cite{kim2015}, which captures more intricate behaviours.
For SHM, a semi-supervised model is preferred to effectively handle unlabeled data. Deep semi-supervised anomaly detection (DeepSAD), proposed by Ruff et al.\ in 2019 \cite{Ruff2019}, extends Deep SVDD by incorporating unlabeled data points. While DeepSAD has proven effective in various non-SHM contexts, its application to HI generation poses challenges. For instance, Han et al.\ \cite{Han2025} used DeepSAD to detect anomalies in vehicle emissions, by setting a threshold anomaly score. They evaluated the performance of models by the area under the receiver operating characteristic curve, thus without considering the distribution of scores on either side of the threshold. This approach, although beneficial for anomaly detection, is inadequate for HI extraction due to the complete neglect of prognostic criteria. Similarly, DeepSAD has been applied with consideration only to the classification of anomalies in hydraulic systems \cite{dong2023}. Other approaches have sought to enhance semi-supervised anomaly detection. Gao et al.\ \cite{Gao2021} developed ConNet, a robust deep anomaly detection model for sparsely labelled data. However, ConNet only considers positive, anomalous, labeled samples, which is unsuitable for HI generation as the healthy state is not considered.}

\textcolor{black}{Recent advancements have specifically targeted HI generation. Frusque et al.\ \cite{Frusque2023} applied DeepSAD to HI generation for rotating machinery, and it was observed that the objective function considers only the norm of the embedding, therefore producing low-rank results with some dimensions not fully utilised. To diversify the representation of degradation trends, they incorporated a diversity term in the loss function, creating Diversity-DeepSAD.
For the PHME2010 milling dataset, Diversity-DeepSAD achieved promising results, with performances of 99\% for $Mo$, 99\% for $Tr$, and 94\% for $Pr$, indicating room for improvement in $Pr$. However, when applied to thermal spray coating monitoring data, the model's performance was lower, with 91\% for $Mo$ and 72\% for $Pr$, while $Tr$ was not reported. Based on the displayed HIs, the $Tr$ score appears low. Notably, the coefficient of variation (CV), useful for comparing variability across datasets, was 24\% for the thermal spray coating dataset. In contrast, aeronautical structures are significantly more challenging, with CV values of 51\% for GW data from five single-stiffener composite panels \cite{Moradi2024} and 87\% for acoustic emission data from 12 specimens \cite{Moradi2023Intelligent, moradi2025novel}. Despite its potential, Diversity-DeepSAD has not yet been applied to aeronautical structures, which present unique challenges for HI construction and remaining useful life (RUL) prediction.}

\textcolor{black}{Moradi \cite{moradi2024designing} developed various semi-supervised frameworks for single-stiffener composite structures. One approach utilised a multi-layer long short-term memory (LSTM) network, employing an intrinsically semi-supervised inductive learning technique, trained on acoustic emission-based time and frequency domain features \cite{Moradi2023Intelligent}. This method achieved a fitness score of 93\% under a rigorous leave-one-out cross-validation (LOOCV) process. However, the fine-tuning process to optimise the number of epochs should be done using only the validation specimen (unit), and another unit, different from the training/validation units, should be investigated to confirm the generalisability of this fine-tuned hyperparameter—something that was not considered in their study. To address this, Moradi et al.\ \cite{moradi2025novel} introduced CEEMDAN-driven semi-supervised ensemble deep learning (CEEMDAN-SSEDL), which combined complete ensemble empirical mode decomposition with adaptive noise (CEEMDAN) for feature extraction and ensemble learning techniques to reduce randomness. By incorporating bidirectional LSTM (BiLSTM) layers during ensemble learning, this approach achieved a fitness score of 91.3\%, while ensuring separate test units were considered.
Moradi et al.'s \cite{Moradi2023Intelligent} semi-supervised inductive learning approach was later applied by Frusque et al.\ \cite{Frusque2023} to the PHME2010 milling dataset, achieving a fitness score of 96\%. However, the details of the architecture and the hyperparameter fine-tuning process were not reported.
Recognising the limitations of history-dependent models, Moradi et al.\ \cite{moradi2024designing, Moradi2023SemiSupverised} noted that such models often perform poorly in the absence of historical data. To overcome this, they introduced the Hilbert transform semi-supervised convolutional neural network (HT-SSCNN), designed to generate HIs from history-independent GW data. Leveraging the GW monitoring technique, this approach achieved a 93\% fitness score. Inspired by this, the methods in this paper are also designed to be history-independent, relying solely on data from the current state of the structure. Given the demonstrated capabilities of anomaly detection algorithms, this study seeks to extend the application of Diversity-DeepSAD to the SHM of aerospace structures.}

\textcolor{black}{Previous applications of DeepSAD, however, have been limited to binary auxiliary labels distinguishing only healthy and failed states. This binary formulation neglects the intermediate degradation states that make up the majority of a structure’s lifetime, and in composites often leaves only a single label at failure, severely restricting the usefulness of labels. To address this gap, we introduce continuous auxiliary labels interpolated across the degradation process, applied only at early-life (e.g.\ 0–25\%) and late-life (e.g.\ 75–100\%) stages, to provide richer supervision and produce smoother HIs.}

\textcolor{black}{Unsupervised models, by design, rely on feedback from the model output to learn without labeled data. A common technique is the autoencoder (AE), which uses a neural network to encode input data into a reduced-dimensional latent space and then reconstruct the input from this representation. Yang et al.\ \cite{yang2018} generated HIs for rotating machinery using sparse AEs, evaluating them based on the Mann-Kendall (MK) monotonicity metric. Lin and Tao \cite{lin2019} extended this approach by employing an ensemble of stacked AEs with linear targets, while Xu and Wang \cite{xu2022} enhanced stacked AEs with noise reduction using an exponential weight moving average model. In 2023, Xu et al.\ \cite{xu2023hi} implemented a deep convolutional AE and Mao et al.\ \cite{mao2023} applied tensor representation. These studies, however, focused on rotating machinery, where the vibrational \textcolor{black}{behaviour} is more self-exposing and sensitive to damage. In contrast, aerospace structures do not exhibit such clear damage effects on the monitored signals, making the application of these methods more challenging.
Variational autoencoder (VAE) architecture is similar to AE, but makes use of a probabilistic latent variable distribution in order to increase generability to data beyond the training set. In 2019, Ping et al.\ \cite{Ping2019} utilised a VAE with a logarithmic normal distribution to generate HIs for turbofan engines (CMAPSS simulated dataset, which is simpler compared to experimental dataset \cite{Moradi2023Intelligent, moradi2025constructing}). In 2020, Hemmer et al.\ \cite{hemmer2020} applied a VAE to rotating machinery at discrete damage levels. While generally increasing with damage level, the HIs do not conform to the prognostic criteria. 
Mitigating this, in 2022, Qin et al.\ \cite{Qin2022} introduced a monotonic degradation constraint in the loss function, resulting in the degradation-trend-constrained VAE (DTC-VAE), which generates HIs that meet the $Mo$ metric. Guo et al.\ \cite{guo2022} proposed a multiscale convolutional AE network for rotating machinery. Further studies in 2024 applied developments of VAE to the generation of HIs for bearings from vibrational data \cite{li2024rotating, li2024vae, milani2024}. However, the application of VAEs for HI generation in aerospace structures remains unattempted. This paper seeks to address this gap by applying the DTC-VAE model to GW signals measured from aerospace structures, aiming to achieve high quality HIs. By leveraging the strengths of both Diversity-DeepSAD and DTC-VAE, the study aspires to advance HI generation for complex aeronautical structures.}

\vspace{10pt}
\section{Method}\label{sec:method}

\noindent This section provides a detailed overview of the methodology adopted in this study, encompassing the experimental setup and the framework for HI generation. Subsection 2.1 introduces the experimental setup which forms the foundation of the analysis. Subsection 2.2 describes the overall proposed two-stage framework for HI generation, followed by Subsection 2.3, which outlines the criteria used to evaluate the generated HIs. Subsection 2.4 delves into the SP techniques applied to preprocess the data, while Subsection 2.5 focuses on the feature extraction process. Finally, Subsection 2.6 presents the feature fusion approach, \textcolor{black}{including} Diversity-DeepSAD, DTC-VAE, and ensemble learning models. \textcolor{black}{Details on code availability are provided in \ref{app: code}.}

\subsection{Experimental setup and data preparation}\label{ss: EXP}
\noindent To evaluate the proposed framework, five single-stiffener composite specimens were subjected to run-to-failure compressive fatigue loading under \textcolor{black}{the ReMAP-H2020 project \cite{Dataverse}}. The specimens consist of IM7/8552 carbon fibre-reinforced epoxy in unidirectional prepreg, formed with layups of ${[\pm45/0/45/90/-45/0]_S}$ and ${[\pm45/0/\pm45]_S}$ \cite{Moradi2024}. The specimens were monitored using different SHM techniques, from which only GWs \textcolor{black}{are used} to construct HIs. 

\begin{figure}[!t]
    \centering
    \includegraphics[width=1\linewidth]{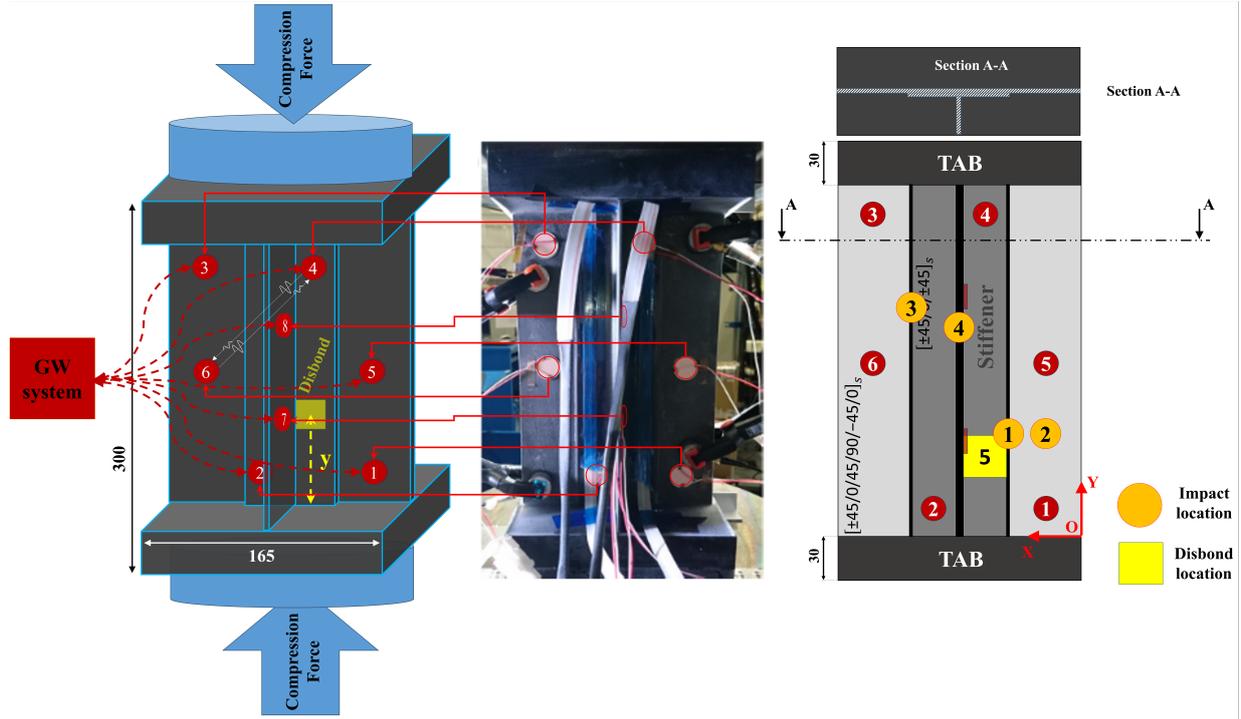}
    \caption{Single T-stiffener CFRP panel under C-C fatigue loading monitored with PZT sensors (red circles).}
    \label{fig:EXPsetup}
\end{figure}

The experimental setup is visualised in \autoref{fig:EXPsetup}\textcolor{black}{, which shows where an impact load (orange circles) or a disbond (yellow squares) was pre-applied to the corresponding numbered specimens}. To carry out GW monitoring, intervals of 5000 cycles were considered \cite{Moradi2024}. Eight PZTs were attached to the structure to function as both actuators and sensors (see red circles in \autoref{fig:EXPsetup}). GW data was collected at frequencies of \SI{50}{\kilo\hertz}, \SI{100}{\kilo\hertz}, \SI{125}{\kilo\hertz}, \SI{150}{\kilo\hertz}, \SI{200}{\kilo\hertz}, and \SI{250}{\kilo\hertz}. At a given moment, one of the PZTs acted as an actuator, while the rest acted as sensors, and this process rotated through all eight PZTs. This created a total of ($8\times7$) 56 actuator-sensor paths. It should be highlighted that the proposed models in this study do not receive indication of uncertainties that may have occurred during testing, such as disbond defects, broken sensors, and impact loading. 

\begin{figure}[!t]
    \centering
    \includegraphics[width=1\linewidth]{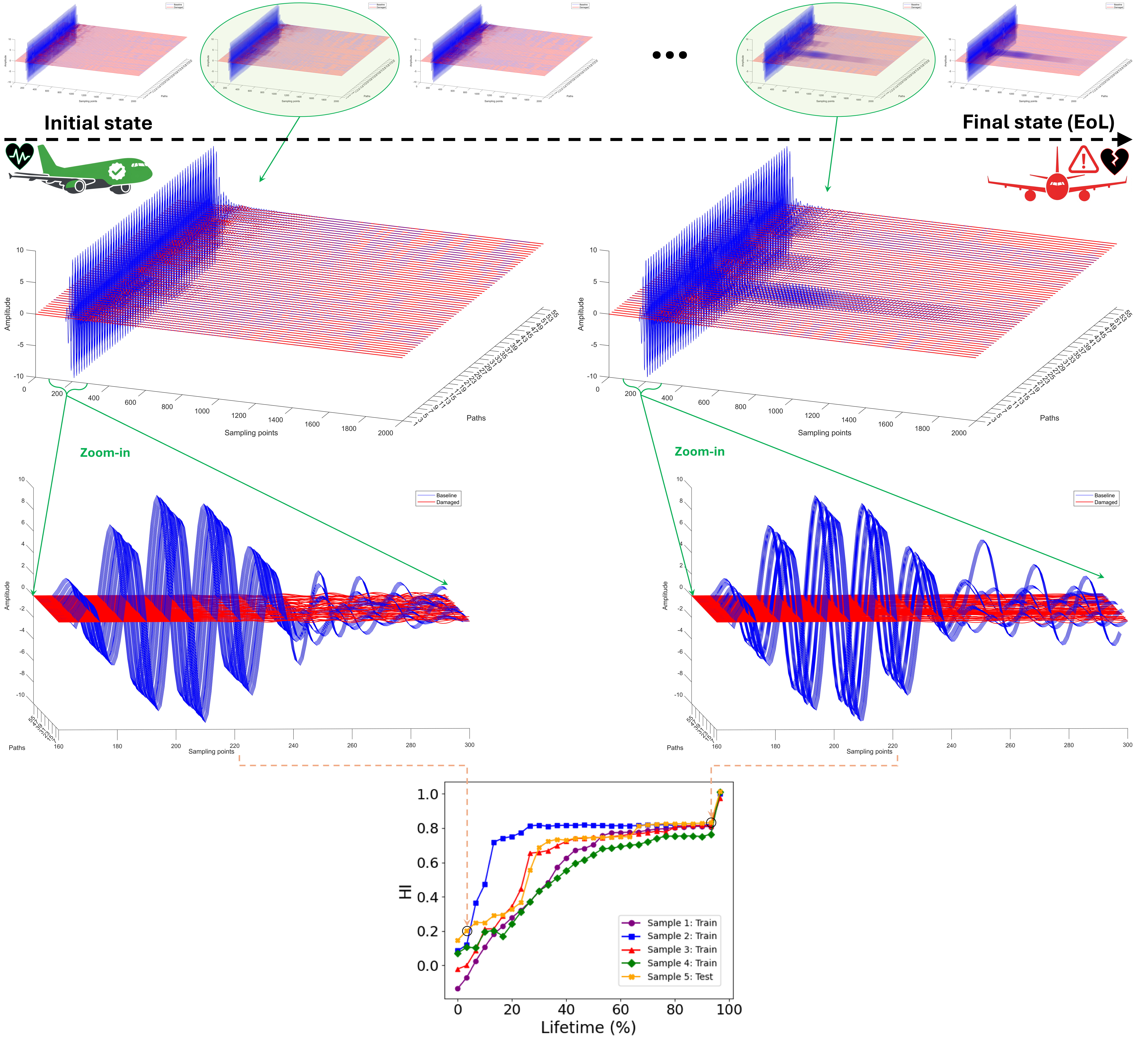}
    \caption{\textcolor{black}{Example initial and end-of-life guided wave signals and corresponding HIs illustrating the structure of the dataset and fatigue-induced waveform evolution.}}
    \label{fig:newfig}
\end{figure}

\textcolor{black}{\autoref{fig:newfig} presents example raw GW signals from the initial (left) and end-of-life (right) states of one composite skin–stiffener panel (composite specimen 5, 125 kHz excitation). Each GW measurement consists of 56 actuator–sensor paths recorded at 6 excitation frequencies, resulting in a 56 × 6 × 2000 data tensor per sample. Zoom-in views highlight changes in waveform characteristics across fatigue progression. The bottom plot shows the corresponding HIs predicted across the lifetimes of multiple units, demonstrating both unit-to-unit variability and complete run-to-failure coverage. In this example, unit 5 was held out entirely for testing, while the remaining units were used for training and validation.}

The input \textcolor{black}{feature data were} arranged in 5 folds, in each of which a different specimen was used for testing while the remaining 4 were for training. \textcolor{black}{This leave-one-unit-out strategy ensures that all measurements from the held-out panel, including its manufacturing tolerances and experimental uncertainties, remain completely unseen during training, providing a stringent cross-unit generalisation test and helping to mitigate overfitting despite the limited number of structural units.}
Training and test data were normalised relative to the training data only using Z-score scaling, in other words, subtracting by the mean and dividing by the standard deviation. 

The model output was also normalised using the training data, with min-max normalisation, to ensure HIs had a value of approximately 0 in the first timestep (healthy state) and approximately 1 at the last (failure).

\subsection{Four-stage framework for HI generation}

\begin{figure}[!tbh]
        \centering
        \includegraphics[width=1\linewidth]{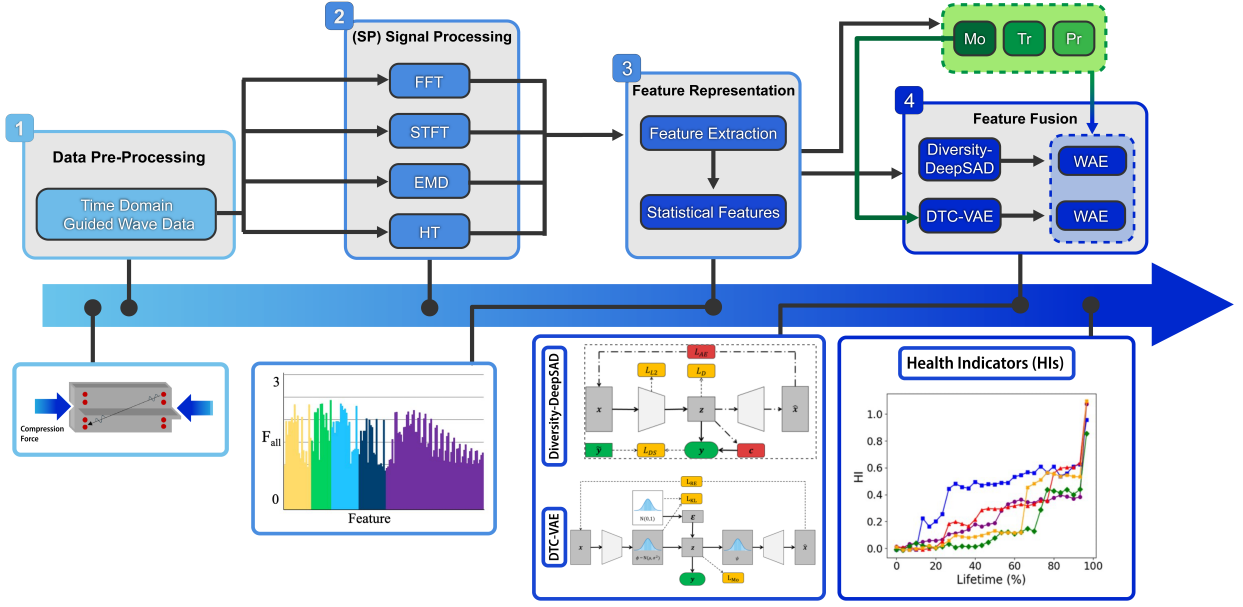}
        
        \caption{Framework for HI generation and evaluation — Step 1: Data Pre-Processing, Step 2: Signal Processing, Step 3: Feature Representation, Step 4: Feature Fusion — with prognostic criteria ($Mo$, $Pr$, $Tr$).}
    \label{fig: FrameworkStage1}
    \end{figure}

\noindent This study followed a four-stage framework, as can be seen in \autoref{fig: FrameworkStage1}, resulting in HI generation through the use of SP techniques and AI models. The first stage was pre-processing data to the correct format. In the second stage, time domain measurements were processed using different SP methods, including fast Fourier transform (FFT), short-time Fourier transform (STFT), empirical mode decomposition (EMD), and Hilbert transform (HT). Thirdly, statistical features were extracted from the processed signals for each path, actuator frequency \textcolor{black}{$f$}, and measurement timestep. As all input features should ideally \textcolor{black}{be} of the same SP method to ensure efficient processing, \textcolor{black}{a single SP method is} used in the workflow. With this in mind, the two SP methods with the highest performing features were selected for further individual use.

In the fourth stage, the features selected were fed into AI models, i.e.\ the Diversity-DeepSAD and DTC-VAE models, where hyperparameters were fine-tuned using Bayesian optimisation. This process was repeated for each combination of excitation frequency of PZTs and SP method. The extracted HIs \textcolor{black}{(i.e.\ $y$)} from each actuating frequency were then fused by an unsupervised ensemble learning model, enabling the evaluation of the remaining combinations of AI model and SP method against the HIs evaluation criteria to determine the optimal final combination.

\subsection{HIs criteria}\label{ss:pc}

\noindent Given the lack of true HI labels, evaluating the quality of the developed HIs necessitates leveraging known physics-based principles and facts. The HI criteria include $Mo$, $Pr$, and $Tr$, and they are standard for this application \cite{moradi2024designing, Moradi2024, Moradi2023Intelligent, Frusque2023, Baptista2022}. \textcolor{black}{While these metrics do not directly measure physical damage, they evaluate whether the learned HI behaves in a physically meaningful and degradation-consistent manner.}

$Mo$ measures the extent to which the HI shows only an increasing or decreasing trend. If maintenance is not performed, health should always decrease with an increasing number of cycles, and thus a monotonic HI should more closely model reality. The modified Mann-Kendall (MMK) formulation of $Mo$ is presented in \autoref{eq:mo} \cite{Moradi2024}:
\small
\begin{equation}\label{eq:mo}
        Mo = \frac{1}{M} \sum^M_{m=1} \left|{\frac{1}{(N_m-1)} \sum^{N_m-1}_{i=1} 
        \frac   {\sum^{N_m}_{j=1, j>i} (t_j - t_i) \cdot sgn(y^{m}_{j}-y^{m}_i)}
                {\sum^{N_m}_{j=1, j>i} (t_j - t_i)}}\right|
\end{equation}
\normalsize

\noindent where $M$ is the number of specimens, $N$ is the number of GW measurements (timesteps), \textcolor{black}{$y^{m}_{i}$} denotes an HI at \textcolor{black}{measurement $i$} of specimen $m$, \textcolor{black}{$t_i$ is similarly the time at the $i^{th}$ measurement, while $j$ can also represent a measurement number in this way.} $sgn(\cdot)$ is the sign function. This formulation of $Mo$ is preferable as it considers time gaps greater than one unit, thus reducing the effect of noise and favoring monotonicity through the whole trend, as opposed to between adjacent points \cite{Moradi2024}.

$Pr$ measures the HIs' deviation at failure between specimens. Preferably, all specimens should have the same value of HI at the end of life, such that they can be reliably used to predict failure. \autoref{eq:pr} expresses $Pr$ \cite{Moradi2024}, where $std(\cdot)$ refers to the standard deviation between specimens.
\begin{equation} \label{eq:pr}
    Pr = exp\left(-\frac{std(y^m_{N})}{\frac{1}{M}\sum^M_{m=1}\left|{y^m_{1}-y^m_{N}}\right|}\right)
\end{equation}

\noindent Finally, $Tr$ is a measure of the correlation between HIs of different units \cite{Moradi2024}. The set of specimens should show the same HI pattern from beginning to end of life, as this makes their behaviour more predictable. $Tr$ is given by \autoref{eq:tr}, where $corr(\cdot)$ is the Pearson correlation function. 
\begin{equation} \label{eq:tr}
    Tr = min_{a,b} \; {corr(y^a,y^b)}, \quad a,b = 1, ..., M
\end{equation}

\noindent \textcolor{black}{In summary, $Mo$ quantifies how consistently the HI evolves monotonically under the assumption that health should not improve without maintenance, $Pr$ measures how closely specimens converge to a common HI value at failure, and $Tr$ evaluates whether specimens follow similar degradation trends throughout their lifetimes.} To effectively evaluate a model's performance on test specimens while mitigating bias towards highly matched training specimens and avoiding the obscuring of unmatched test specimen \textcolor{black}{behaviour}, alternative evaluation criteria must be established. These criteria should prioritise the \textcolor{black}{behaviour} of test specimens over those used during training, ensuring a more reliable assessment \cite{Moradi2024, moradi2025novel}. These criteria include $Mo_{test}$, which evaluates the monotonicity of only the test specimen(s), and $Pr_{test}$, which assesses the prognosability by emphasizing the deviation of the test specimen(s) relative to those used for training. These metrics are  defined in \autoref{eq:mo_test} and \autoref{eq:pr_test} \cite{Moradi2024}:

\begin{equation} \label{eq:mo_test}
    Mo_{test} = \left|{\frac{1}{(N_{test}-1)} \sum^{N_{test}-1}_{i=1} 
        \frac   {\sum^{N_{test}}_{j=1, j>i} (t_j - t_i) \cdot sgn(y^{m}_{j}-y^{m}_i)}
                {\sum^{N_{test}}_{j=1, j>i} (t_j - t_i)}}\right|
\end{equation}

\begin{equation} \label{eq:pr_test}
    Pr_{test} = exp\left(-\frac {\left| {y^{test}_N -  \frac{1}{M^{train}}\sum^{M^{train}}_{m=1} y^m_N } \right|}
                                {\frac{1}{M} \sum^M_{j=1} \left| y_1^j - y_N^j \right|}\right)
\end{equation}

\noindent where the index $train$ or $test$ represents train or test specimens, respectively, and thus $M^{train}$ refers to the number of the training specimens.

The fitness score $F_{all}$ is defined by \autoref{eq:fitness}, and measures the overall quality of HIs \cite{Moradi2024}, while $F_{test}$ is defined similarly in \autoref{eq:ftest} with a focus on test units. In order to weight the criteria equally, the control constants $k_{Mo}$, $k_{Pr}$, and $k_{Tr}$ are set to equal 1. Both fitness scores $F_{all}$ and $F_{test}$ are calculated and discussed in this study.
\begin{equation}\label{eq:fitness}
    F_{all} = k_{Mo} \cdot Mo + k_{Pr} \cdot Pr + k_{Tr} \cdot Tr
\end{equation}
\begin{equation}\label{eq:ftest}
    F_{test} = k_{Mo} \cdot Mo_{test} + k_{Pr} \cdot Pr_{test} + k_{Tr} \cdot Tr
\end{equation}

\noindent \textcolor{black}{The prognostic criteria $Mo$, $Pr$, and $Tr$ each take values in the range $[0,1]$, where 0 represents the worst behaviour and 1 represents the ideal expected behaviour of a degradation trajectory. The composite fitness scores $F_{all}$ and $F_{test}$ therefore lie in the range $[0,3]$, with 3 indicating the best overall performance across the three prognostic criteria. When reporting performance as a percentage (e.g.\ 92.3\%), this corresponds to the fitness score normalised by its maximum value of 3.}

\subsection{Signal processing}\label{ss:sp}

\noindent SP techniques were applied to transform the measured signals into the frequency and time-frequency domains, in order to highlight relevant information, such as frequency components or noise isolation. These transformations enabled the extraction of relevant features for further analysis. The techniques used include FFT, STFT, EMD, and HT.

The FFT algorithm transforms the original temporal signal into the frequency domain by processing the entire time domain at once and outputting corresponding amplitudes for the underlying frequencies. However, it does not retain temporal variations in the signal, such as how these frequencies evolve over time, which may reveal additional information about the structure's health. In contrast, STFT preserves time-frequency information by dividing the data into smaller time intervals and analyzing localised segments, making it advantageous for non-stationary signals where frequency content changes over time. The result of STFT is a 2D matrix, where time and frequency are the axes, and amplitude is the value at each position. Nevertheless, there is a nonlinear trade-off between time and frequency resolution: too many time segments reduces the available frequency samples, and vice versa. In this study, it was found that performing STFT with an FFT length of 250 maximised the number of data points, resulting in an overlap length of 125.

In the case of EMD, the sifting algorithm was implemented to iteratively decompose the signal into intrinsic mode functions (IMFs) in the time domain. The process involves constructing upper and lower envelopes of the signal using cubic splines to interpolate local maxima and minima, respectively. The mean of these envelopes is then subtracted from the signal at each iteration, refining the signal until it satisfies the conditions for an IMF. Suitable stopping conditions are necessary to avoid valuable information being lost \cite{Boudraa2007}. Through trial and error, these criteria were set to a minimum standard deviation of $0.1$  and a maximum of 10 iterations.

The HT algorithm produces a complex-valued signal, the analytic signal, from a real signal. The real component represents the original signal, while the imaginary component is created by phase-shifting each frequency component by $\frac{\pi}{2}$ \cite{Feldman2011}. The analytic signal therefore contains information concerning instantaneous amplitude and phase. 

By employing these SP techniques, features can be extracted that capture essential signal characteristics across the time, frequency, and time-frequency domains, forming a robust foundation for subsequent analysis.

\subsection{Feature extraction}\label{ss:features}
\noindent Statistical features, as detailed in \ref{app:A}, were extracted from the time, frequency, and time-frequency domains to characterise the signals comprehensively. \textcolor{black}{The selected feature set aligns with established approaches in the literature \cite{Moradi2023Intelligent}. They were chosen due to their proven capability to capture diverse and informative aspects of signal behaviour.}

From the time domain, 19 statistical features were extracted, as shown in \autoref{tab:timedomainfeatures}. These features were computed for the original time-domain signals as well as the processed signals using HT and EMD. 
From the frequency domain, 14 statistical features were extracted from the FFT-transformed signals, as listed in \autoref{tab:freqdomainfeatures}. 
For the time-frequency domain, the output of STFT was divided into 17 time windows, each with discrete frequency components. To reduce dimensionality, four statistical measures—mean, standard deviation, skewness, and kurtosis—were calculated across the frequency axis within each time window. This resulted in a total of 68 features (4 measures × 17 time windows), as summarised in \autoref{tab:STFTdomain}.

The extracted features are outlined in \autoref{tab:featureslayout}. By systematically extracting statistical features across multiple domains, the framework ensures a concise yet rich representation of the signal information.

\begin{table}[!tbh]
    \centering
    \caption{Feature numbers extracted from the results of each SP method.}
    \vspace{-5pt}
    \begin{tabular}{lccccccc}\hline
          \textbf{Method} & \textbf{Raw Data} & \textbf{FFT} & \textbf{HT} & \textbf{EMD} & \textbf{STFT} \\\hline
         \textbf{Features domain} & Time & Frequency & Time  & Time & Time-Frequency \\
          \textbf{Features} & 1-19 & 20-33 & 34-52 & 53-71 & 72-139   \\\hline
    \end{tabular}
    \label{tab:featureslayout}
\end{table}

For the first stage, features across different actuator-sensor paths and frequencies are averaged, resulting in 139 features per timestep (GW measurement) for each specimen.  \textcolor{black}{ This averaging is intended to capture a robust representation of the specimen's response, to emphasize consistent trends across multiple paths and frequencies whilst reducing the influence of local variations or noise.} After obtaining all timesteps and specimens, the HI prognostic criteria defined in Section~\ref{ss:pc} are applied to evaluate each feature. \textcolor{black}{In order} to avoid anomalously poor features disproportionately impacting the SP method selected, the mean fitness score across all features is calculated to serve as a benchmark. Features with fitness scores exceeding this benchmark are retained for further analysis. \textcolor{black}{This benchmark based filtering process acts as a feature selection step, comparable to Baraldi et al.\ \cite{Baraldi2018OptimizationHI} method of identifying the most informative set of features to ensure that only consistently relevant features contribute to HI development.} The SP method yielding features with the highest remaining mean fitness scores is identified as optimal for developing an efficient framework to extract HI, \textcolor{black}{providing a systematic approach for feature selection rather then relying on arbitrary choices.}

\subsection{Feature fusion}\label{ss:featurefusion} 
\noindent Given the extracted and selected features as inputs, two deep learning models were developed to generate HIs, i.e.\ semi-supervised Diversity-DeepSAD and unsupervised DTC-VAE, which will be presented in this section.

\subsubsection{Diversity-DeepSAD}

\noindent The original DeepSAD model incorporates labeled training data in the loss function of the otherwise unsupervised Deep SVDD model \cite{Ruff2019}. This approach transforms input data into a hyperspace of reduced dimensions, in which the distance to a point, the hypersphere \textcolor{black}{centre} $\mathbf{c}$, represents the extent to which the datapoint is an anomaly.
Frusque et al.\ \cite{Frusque2023} modified the loss function further to include a term to increase diversity, resulting in Diversity-DeepSAD for HI generation. This was demonstrated on a rotational machinery dataset with hundreds of measurement timesteps, a condition not often met in many other applications. For example, the ReMAP dataset may contain as few as 30 timesteps before failure. Consequently, it is necessary to explore the impact of alternative functions for generating auxiliary artificial labels and to evaluate the influence of the number of labeled measurements.

In prior studies \cite{Ruff2019, Frusque2023}, a binary set of auxiliary labels $\tilde{y}_j \in \{1, -1\}$ was employed, where a label of 1 was assigned to measurements representing a very healthy condition, and -1 corresponded to those reflecting a severely damaged state. This approach focuses solely on two extreme conditions. However, these two phases typically occupy only a small portion of the system's entire lifetime, limiting the potential to fully leverage auxiliary labels. This issue is particularly pronounced in the severely damaged phase, where measurements are rare, often with only a single observation available at the point of failure for assigning the label -1. To address these limitations, this study generated continuous auxiliary labels $\tilde{y}_j \in [1, -1]$. \textcolor{black}{Although many functions were considered, best performance was achieved through a simpler labeling scheme} according to \autoref{eqn:labels}:

\begin{equation}\label{eqn:labels}    
    \tilde{y}_j = 1-2\frac{t_j}{t_N}
\end{equation}

\begin{figure}[htbp]
\includegraphics[width=0.9\linewidth]{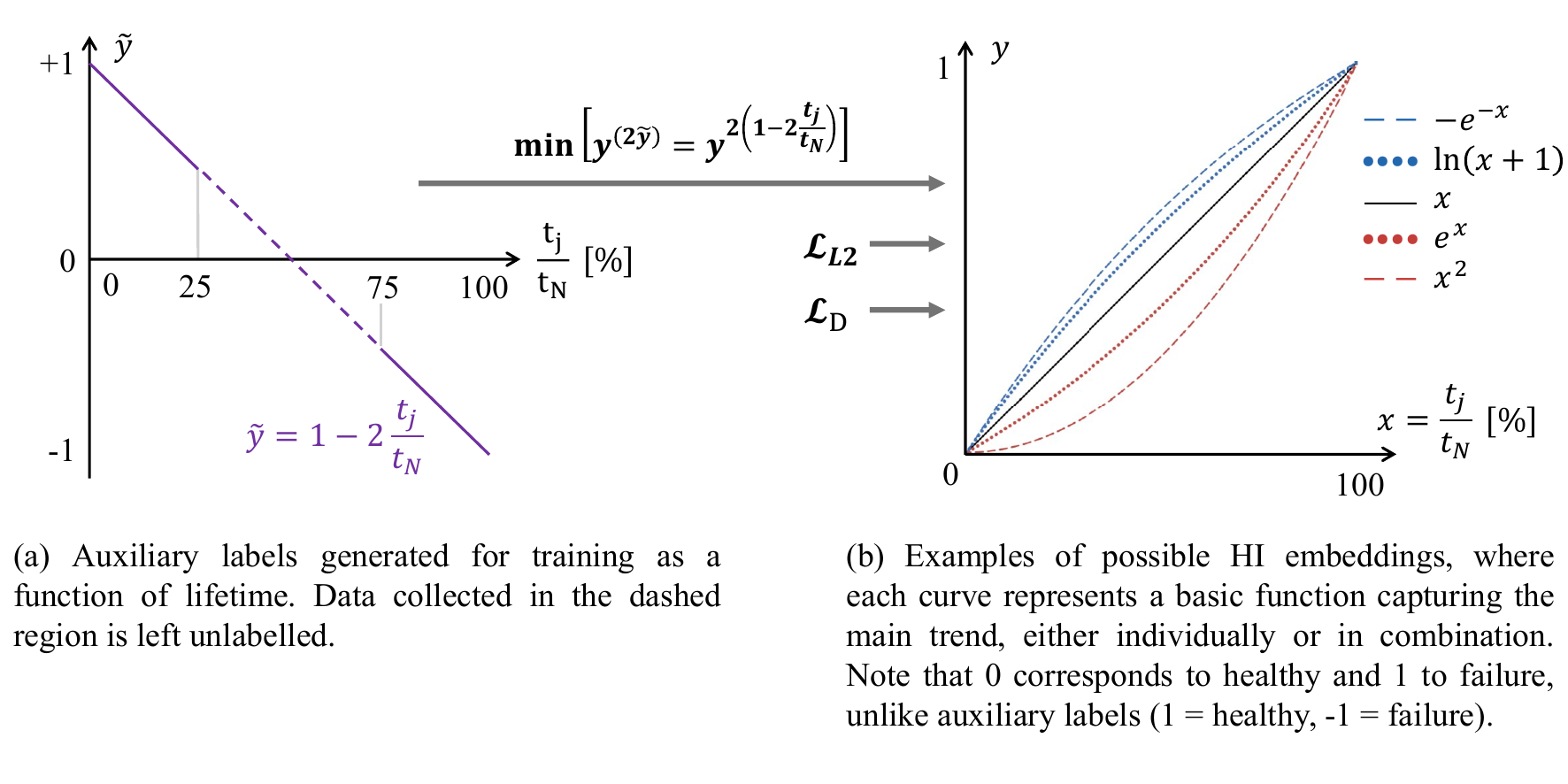}
\caption{\textcolor{black}{Relationship between new auxiliary labels and embedding output expected for the proposed Diversity-DeepSAD.}}
\label{fig:ddGraphs}
\end{figure}

\noindent \textcolor{black}{where a linear function} was applied for the first and last quarters of the lifetime. \textcolor{black}{The use of a linear function avoids imposing any assumed physical degradation which could reduce the flexibility of the model, given this is an extension from a pure binary $\pm1$ formulation.} The quarter ratio was established experimentally to produce improved results. As visualised in \hyperref[fig:ddGraphs]{Figure 4a}, the remaining data points do not leverage auxiliary labels, providing flexibility to the model, given the uncertainty of the true health status for these intermediate conditions.

It should be noted that auxiliary labels $\tilde{y}_j$ differ from the expected \textcolor{black}{HI}, denoted by $y_j$. \textcolor{black}{\autoref{eqn:labels} therefore need not bear resemblance to the physical behaviour of the sample.} Assuming the hypersphere \textcolor{black}{centre} $\mathbf{c}$ represents the healthy state, the HI is defined as the distance from that \textcolor{black}{centre}:

\begin{equation}\label{eqn:embed0}
    \mathrm{HI} = y_j=||z_j-\mathbf{c}||    
 \end{equation}

\noindent where $z_j$ denotes the latent space representation obtained from the autoencoder. When $z_j$ is close to the healthy state $\mathbf{c}$, $y_j$ approaches 0, indicating minimal deviation from the healthy condition. Conversely, as $z_j$ moves farther from $\mathbf{c}$, $y_j$ increases, reflecting a more degraded condition. In this framework, the maximum value of 1 is assigned to $y_j$ in the fully damaged state. Consequently, the HI can freely exhibit linear or non-linear \textcolor{black}{behaviour}\textemdash with different main trends, such as exponential or polynomial \textcolor{black}{behaviour}\textemdash between these bounds, as illustrated in \hyperref[fig:ddGraphs]{Figure 4b}. 

Denoting the encoder function as $\phi$, the HI can be expressed as:

\begin{equation}\label{eqn:embed}
    y_j=||\phi_\theta (\mathbf{x}_j)-\mathbf{c}||
 \end{equation}

\noindent where $\phi_\theta (\mathbf{x}_j)$ represents the latent space embedding of the input data $\mathbf{x}_j$ at timestep $j$, and $\theta$ denotes the learnable parameters of the encoder.
The sum of squares of these embeddings, raised to the power of the auxiliary label $\tilde{y}_j$ where applicable, is minimised using the following loss function:

\begin{equation} \label{eq:l_ds}
    \mathcal{L}_{DS} = \frac{1}{n+u}\sum_{i=1}^u \left(y_i\right)^2  + \frac{\eta}{n+u}\sum_{j=1}^n\left(y_j+\epsilon\right)^{2\tilde{y}_j}
\end{equation}

\noindent in which $n$ represents the number of labeled samples, $u$ is the number of unlabeled samples, and $\eta$ is a hyperparameter that controls the contribution of the auxiliary labels. To prevent numerical zero errors when $\tilde{y}_j$ is negative, a small term $\epsilon$ is added to labeled HIs. This loss component is the first term in the overall Diversity-DeepSAD loss function:

\begin{equation}\label{eqn:deepsadloss}
    \mathcal{L}_{Diversity-DeepSAD} = \mathcal{L}_{DS} + \nu \mathcal{L}_{L2} + \lambda \mathcal{L}_D
\end{equation}

\noindent where \textcolor{black}{$\mathcal{L}_{L2}$ represents $L_2$ regularisation \cite{gauss1809}, minimising the sum of squares of node weights to encourage a greater distribution of smaller weights and prevent overfitting.} $\nu$ and $\lambda$ are hyperparameters controlling the weights of the \textcolor{black}{regularisation} terms $\mathcal{L}_{L2}$ and $\mathcal{L}_{D}$, respectively. \textcolor{black}{$\mathcal{L}_{D}$ is} defined as follows:

\begin{equation}
    \mathcal{L}_{D} = \sum_{h=1}^H(\zeta_h-\ln(\zeta_h))
\end{equation}

\noindent where $\zeta_{h}$ represents the $h^{\text{th}}$ \textcolor{black}{out of $H$ eigenvalues} of the Gram matrix $G=Z^TZ$, in which $Z$ is the DeepSAD embedding. $\mathcal{L}_{D}$ promotes a greater distribution of training data in the model output \cite{Frusque2023}.
These three terms therefore work together to ensure that the model effectively learns the latent representations, while the \textcolor{black}{regularisation} terms help prevent overfitting and promote diversity within the learned embeddings.

A standard AE is always used to pretrain the network weights (learnable parameters $\theta$). This AE is formed by adding a decoder $\psi$ which inverts the Diversity-DeepSAD network layers, considering the mean squared difference between input and output data as a loss function, $\mathcal{L}_{AE}$. This loss function is used to pre-train the autoencoder, but is disregarded in the training of Diversity-DeepSAD in \autoref{eqn:deepsadloss}. The hypersphere \textcolor{black}{centre} \( \mathbf{c} \) is initialised by computing the average output from the model during a forward pass on the training data. Initially, \( \mathbf{c} \) is set to a zero matrix. In the pretraining phase, denoted by \( \phi_{\theta}^{[0]} \), the model performs a forward pass on the training data, and the outputs for each sample are accumulated. The number of samples is incremented with each data point, and the cumulative outputs are added to \( \mathbf{c} \). After processing all the data, the matrix \( \mathbf{c} \) is normalised by the total number of samples to obtain the average. To avoid numerical issues, any component of \( \mathbf{c} \) with an absolute value smaller than \( \epsilon \) is adjusted as follows:

\begin{equation}
    \mathbf{c}=\begin{cases} \phi_{\theta}^{[0]} + \epsilon & \text{ if } \;\;\;\;\; 0<\phi_{\theta}^{[0]}<\epsilon \\ \phi_{\theta}^{[0]} - \epsilon & \text{ if } \; -\epsilon<\phi_{\theta}^{[0]}<0 \end{cases}
\end{equation}

\begin{figure}[!b]
        \centering
        \vspace{-10pt}
        \includegraphics[width=1\linewidth]{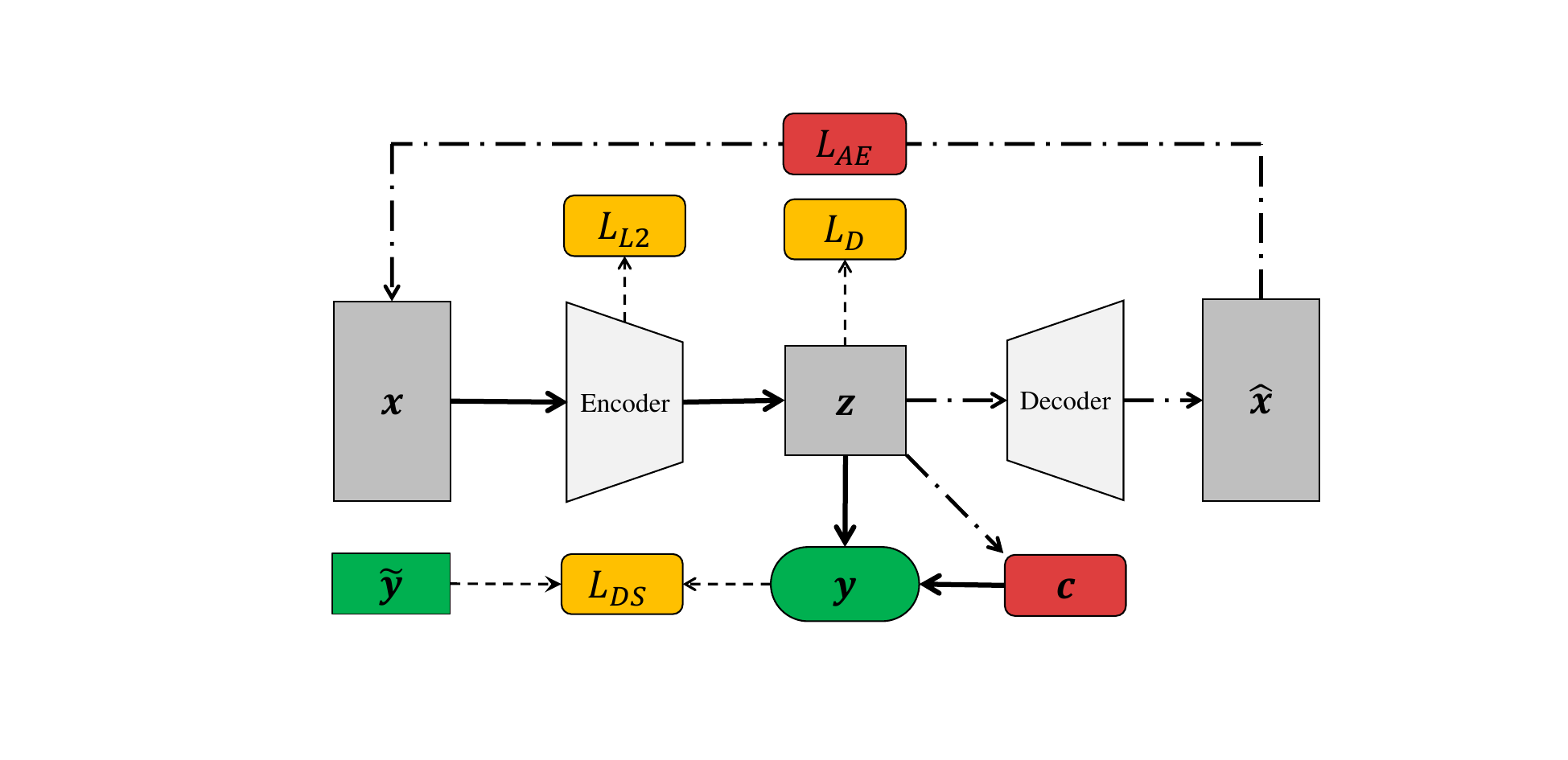}
        \vspace{-55pt}
        \caption{Architecture of Diversity-DeepSAD. Dash-dotted lines indicate the pretraining autoencoder and dashed lines indicate the loss function.}
        \label{fig:DeepSAD_architecture}
\end{figure}

\noindent This adjustment ensures stability in the hypersphere \textcolor{black}{centre} values, helping the model to establish a reference point for the healthy state. The hypersphere \textcolor{black}{centre} $\mathbf{c}$ serves as a critical foundation for anomaly detection and \textcolor{black}{HI} estimation.
The model architecture is illustrated in \autoref{fig:DeepSAD_architecture}, where red, yellow, and green represent pretraining, training, and model output, respectively.

\textbf{Diversity-DeepSAD architecture and hyperparameter optimisation:} \textcolor{black}{The Diversity-DeepSAD model was implemented with 6 hidden layers, each containing half the number of neurons as the previous layer, and leaky rectified linear units (Leaky ReLU) used as activation functions between layers. The hyperspace was set to be 16-dimensional (resulting in \( \mathbf{c} \) with dimensions \( 4 \times 4 \)), which was found to enhance the performance of AE during pretraining.}

\textcolor{black}{Hyperparameters with the strongest influence on model performance, i.e.\ batch size, learning rate (AE and DeepSAD stages) and the number of training epochs, were optimised via Bayesian optimisation. Optimisation was performed independently for every pair of frequency and feature type. Each run comprised 20 objective evaluations: the first 10 corresponded to uniform random sampling of the search space, and the remaining 10 were selected by a Gaussian-process surrogate model. The surrogate used a Matérn kernel with automatic relevance determination of the length scales, and a default acquisition strategy, which adaptively balances expected improvement, probability of improvement and lower confidence bound. The scalar optimisation objective was the $F_{all}$ score.}

\textcolor{black}{At every Bayesian optimisation iteration, a DeepSAD model was trained on the four training specimens of the fold, and $F_{all}$ was computed using only these units. The held-out specimen was excluded entirely during optimisation and used solely for testing. Across frequencies, optimisation trajectories typically stabilised after 15-18 evaluations, with flat incumbent curves and repeated sampling of similar hyperparameters, indicating convergence. For each (frequency, method), the hyperparameter set achieving the maximum observed $F_{all}$ was selected and retrained once on the full training set prior to evaluation on the test specimen.}

\textcolor{black}{All other hyperparameters were fixed based on literature \cite{Frusque2023}: $\nu=10$, $\eta=10$, $\lambda=0.001$ and $\epsilon=1\cdot10^{-6}$. Section~\ref{subsec:sensitivity_method}, Section~\ref{subsec:sensitivity_results} and \ref{app: sensitivity} discuss the sensitivity analysis to demonstrate that these fixed weights lie within broad regions of high performance, confirming their robustness.}

\begin{table}[!tbh]
    \centering
    \caption{Diversity-DeepSAD hyperparameters space.}
    \vspace{-5pt}
    \begin{tabular}{lcc}\hline
          \textbf{Hyperparameter} & \textbf{Min value} & \textbf{Max value} \\\hline
          Batch Size & 50 & 150 \\
          Learning Rate (Pretraining)& 0.0001 & 0.001 \\
          Learning Rate & 0.0001 & 0.001 \\
          Number of Epochs (Pretraining) & 5 & 20 \\
          Number of Epochs & 50 & 200 \\\hline
    \end{tabular}
    \label{tab:hypspace_Diversity-DeepSAD}
\end{table}

\subsubsection{DTC-VAE}
\noindent VAEs are probabilistic generative models that encode an input vector $x_j$ into a lower-dimensional latent space $z_j$. While traditional autoencoders map $x_j$ to $z_j$ directly, VAEs instead learn a probabilistic mapping $\phi (z_j|x_j)$ in the encoder. This is parametrised by a mean $\mu$ and variance $\sigma^2$, allowing the model to represent uncertainty in the latent space. The latent variable, which in this paper is considered as the HI, is sampled according to \autoref{eq:z_vae}, where $\varepsilon$ is Gaussian noise added for variability.
\begin{equation} \label{eq:z_vae}
    z = \mu + \sigma \odot \varepsilon, \quad  \varepsilon \sim \mathcal{N}(0, 1)
\end{equation}

\begin{figure}[!b]
        \centering
        \includegraphics[width=1\linewidth]{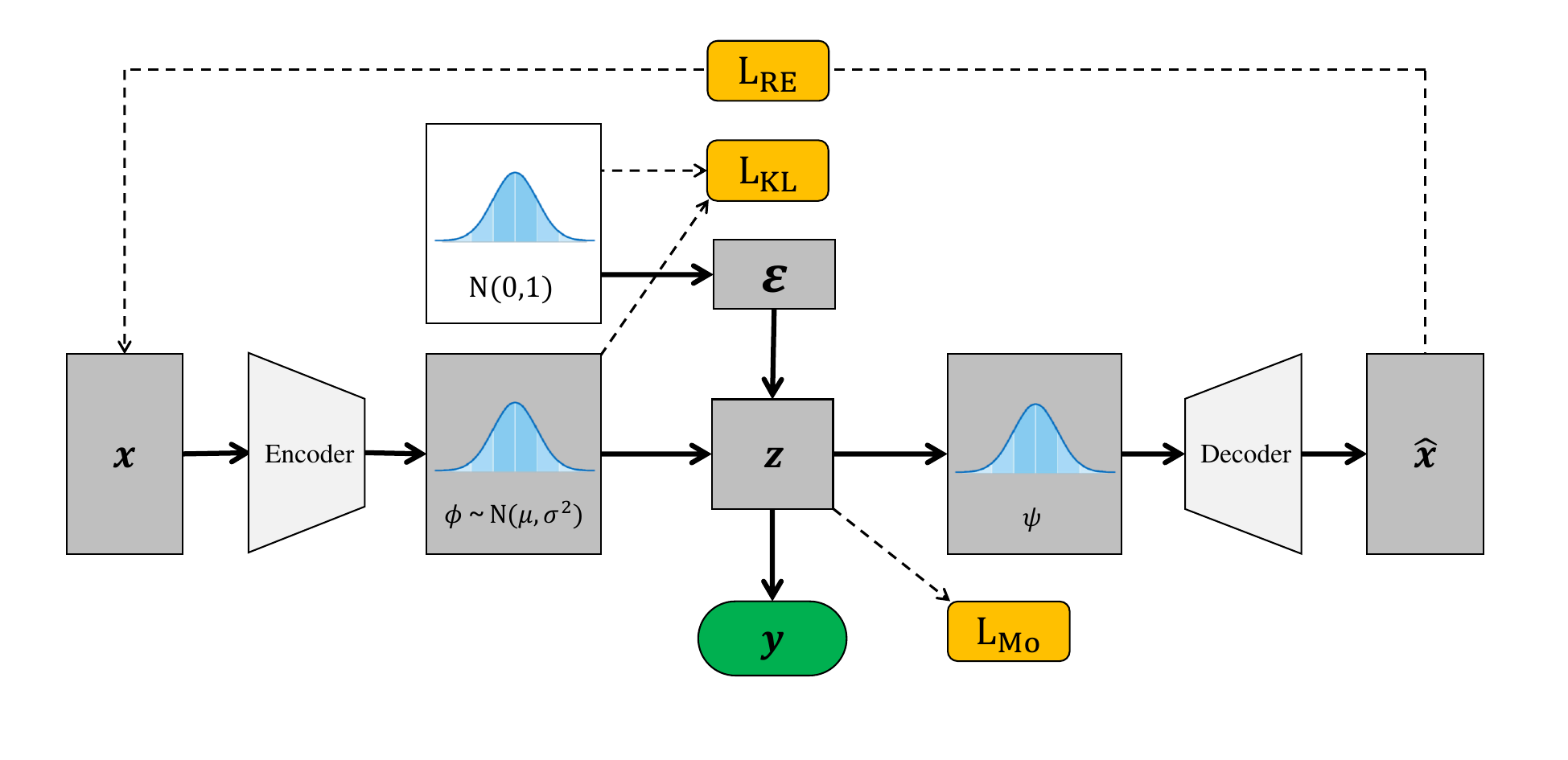}
        \vspace{-30pt}
        \caption{Architecture of degradation-trend-constrained variational autoencoder (DTC-VAE).}
        \label{fig:VAE}
\end{figure}

Finally, the decoder produces a reconstruction of the input $\hat{x}_j$ from the latent variable $z_j$, using the decoder probability distribution $\psi (x_j|z_j)$. This mechanism is illustrated in \autoref{fig:VAE}, where yellow and green represent training and model outputs, respectively.

The DTC-VAE model is optimised by minimising the loss function, $\mathcal{L}_{DTC-VAE}$, which is a summation of the Kullback-Leibler divergence ($\mathcal{L}_{KL}$) \textcolor{black}{\cite{kingma2014autoencoding}}, reconstruction loss ($\mathcal{L}_{RE}$) and monotonicity constraint loss ($\mathcal{L}_{Mo}$):

\begin{equation}\label{eq:vae_loss_function}
    \mathcal{L}_{DTC-VAE} = \alpha \mathcal{L}_{KL}(\phi, \psi) + \beta \mathcal{L}_{RE}(\phi, \psi) + \gamma \mathcal{L}_{Mo}(z)
\end{equation}

\noindent where $\alpha$, $\beta$, and $\gamma$ are their respective \textcolor{black}{hyperparameter weights}.

Secondly, $\mathcal{L}_{RE}$, portrayed in \autoref{eq:RE_loss}, characterises the difference between the encoder input $\mathbf{x}_j^m$ and reconstructed decoder output $\hat{x}_j^m$ for data of a specimen $m$ and timestep $j$, ensuring that these two are as similar as possible.

\begin{equation}\label{eq:RE_loss}
    \mathcal{L}_{RE}(\phi, \psi) = \sum_{m=1}^{M}\sum_{j=1}^{N_m} (\mathbf{x}_j^m - \mathbf{\hat{x}}_j^m)^2
\end{equation}

Finally, the latent variable is prompted to follow a monotonic trend by adding the monotonicity constraint loss, $\mathcal{L}_{Mo}$. This was proposed by Qin et al.\ \cite{Qin2022}, ensuring the extracted HIs followed a monotonic trend by adding a degradation-trend constraint, hence the name DTC-VAE. While the other loss terms are common for VAE architectures, this one is unique to the DTC-VAE model and has a direct effect on the fitness of the output. The monotonicity constraint loss is expressed as follows:

\begin{equation}\label{eq:Mo_loss}
    \mathcal{L}_{Mo}(z) = \sum_{j=2}^N (z_j^m - z_{j-1}^m - r)^2
\end{equation}

\noindent where $z_j^m$ represents the latent variable for specimen $m$ at timestep $j$, while the rate of degradation $r$ is a hyperparameter ensuring that $z_j^m > \left(z_{j-1}^m + r\right)$, which controls the magnitude of change of an HI between timesteps. Its value was determined from the original DTC-VAE study \cite{Qin2022}, in essence, a randomly selected constant in the range $(9, 10)$.

The original HIs criteria functions could not be adapted for gradient descent, given the fact that they are not differentiable and require comparison of latent outputs across specimens, inaccessible during model training. In the case of monotonicity, originally presented in \autoref{eq:mo}, an alternate formulation $\mathcal{L}_{Mo}$ has been implemented, which ensures high monotonicity scores and is compatible with the training process. Ideally, all prognostic criteria ($Mo$, $Pr$, $Tr$) shall be implemented into the loss function in future to maximise fitness.

A common issue in VAEs is posterior collapse, where the latent variables fail to encode meaningful information because the decoder learns to reconstruct inputs directly and the posterior distribution collapses to the prior. This typically occurs when the KL divergence term dominates the loss \cite{Lucas2019}. In this work, the risk of collapse is mitigated through optimisation of the loss weights $\alpha$, $\beta$, and $\gamma$, following Qin et al.\ \cite{Qin2022}, and by the inclusion of the monotonicity constraint $\mathcal{L}_{Mo}$, which forces the latent variables to capture degradation trends instead of defaulting to the prior.

\textbf{DTC-VAE architecture and hyperparameter optimisation:} The DTC-VAE model consists of 3 components: the encoder, latent space and decoder. The encoder's only hidden layer uses a sigmoid activation function and has a varying number of neurons, kept as a hyperparameter. This is followed by two output layers of size 1, representing the mean and log variance of the latent space. As explained by \autoref{eq:z_vae}, the \textcolor{black}{reparameterisation} trick is used to allow for gradient calculation and thus backpropagation. The decoder acts as a mirror to the encoder, it takes the latent variable as input to a hidden layer with sigmoid activation functions. This hidden layer has the same number of neurons as in the encoder. It is followed by an output layer the same size as the input with linear activation function, which returns the reconstructed input. All weights are initialised with uniform Glorot (Xavier) initialisation \cite{Xavier} due to its state-of-the-art performance with sigmoid activation functions and improved stability. Adam is used as the optimiser.

\textcolor{black}{The hyperparameters were optimised over the space detailed in \autoref{tab:hypspace_VAE}, again using Bayesian optimisation with Gaussian processes and $F_{all}$ as the objective. For each frequency and feature type (SP method), the optimiser was run for 40 evaluations: the first 10 corresponded to random samples from the hyperparameter space, and the remaining 30 were guided by the Gaussian-process surrogate. At each evaluation, DTC-VAE was trained on the four training specimens of the current fold, and $F_{all}$ was computed from the resulting HIs of these four specimens only. The held-out panel did not contribute to the optimisation objective.}

\textcolor{black}{Convergence behaviour was similar to that of Diversity-DeepSAD: the best observed $F_{all}$ typically stabilised after approximately 15 iterations, with subsequent iterations exploring hyperparameters within a narrow band of performance. The final configuration (per frequency and method) was selected as the hyperparameter set achieving the highest $F_{all}$ across all iterations and retrained on the full training set before computing the test metrics. The sensitivity analysis in Section~\ref{subsec:sensitivity_method}, Section~\autoref{subsec:sensitivity_results} and \ref{app: sensitivity} confirms that the selected loss-weight combinations lie inside broad plateaus of high performance, rather than isolated optima.}

\begin{table}[!tbh]
    \centering
    \caption{DTC-VAE hyperparameters space.}
    \vspace{-5pt}
    \begin{tabular}{lcc}\hline
          \textbf{Hyperparameter} & \textbf{Min value} & \textbf{Max value} \\\hline
          Hidden Layer Size & 40 & 60 \\
          Batch Size & 75 & 95 \\
          Learning Rate & 0.001 & 0.01 \\
          Number of Epochs & 500 & 600 \\
          $\alpha$ & 1.4 & 1.8 \\
          $\beta$ & 0.05 & 0.1 \\
          $\gamma$ & 2.6 & 3 \\\hline
    \end{tabular}
    \label{tab:hypspace_VAE}
\end{table}

\subsubsection{Ensemble learning model}

\noindent To ensure the reproducibility of results, it is essential to initialise the models with predetermined random seeds. After hyperparameter optimisation with a given seed, the models were each trained for 5 different random seeds to evaluate the stability of the models. The HIs generated by each model were then mean-averaged, and the HIs criteria applied to obtain the fitness scores reported in \autoref{sec:results}.

The weighted averaging ensemble (WAE) model was implemented to fuse the HI results of the \textcolor{black}{$N_f$} different excitation GW frequencies. This is done to improve performance by reducing instability, and removing the need to select and rely on only one frequency. It was chosen to use the normalised fitness scores of the averaged HIs as weights $\tilde{\omega}_f$ for the same frequency, as this method was most successfully implemented previously \cite{Moradi2024}. This method is described by \autoref{eqn:WAE}, with the constants $\omega_f$ adjusted depending on their fitness, producing a new weighted-average HI value $\Bar{y}$.

\begin{subequations} \label{eqn:WAE}
\textcolor{black}{
\begin{equation}
    \bar{y} = \sum_{f=1}^{N_f}\tilde{\omega}_f \bar{y}^{f}
\end{equation}}
\begin{equation}
    \tilde{\omega}_f = \frac{\omega_f}{\Sigma_{f=1}^F \omega_f}
\end{equation}
\begin{equation}
    \omega_f = F_{all}\left(\bar{y}^{f}\right)
\end{equation}
\end{subequations}

\begin{figure}[!b]
    \centering
    \includegraphics[width=1\linewidth]{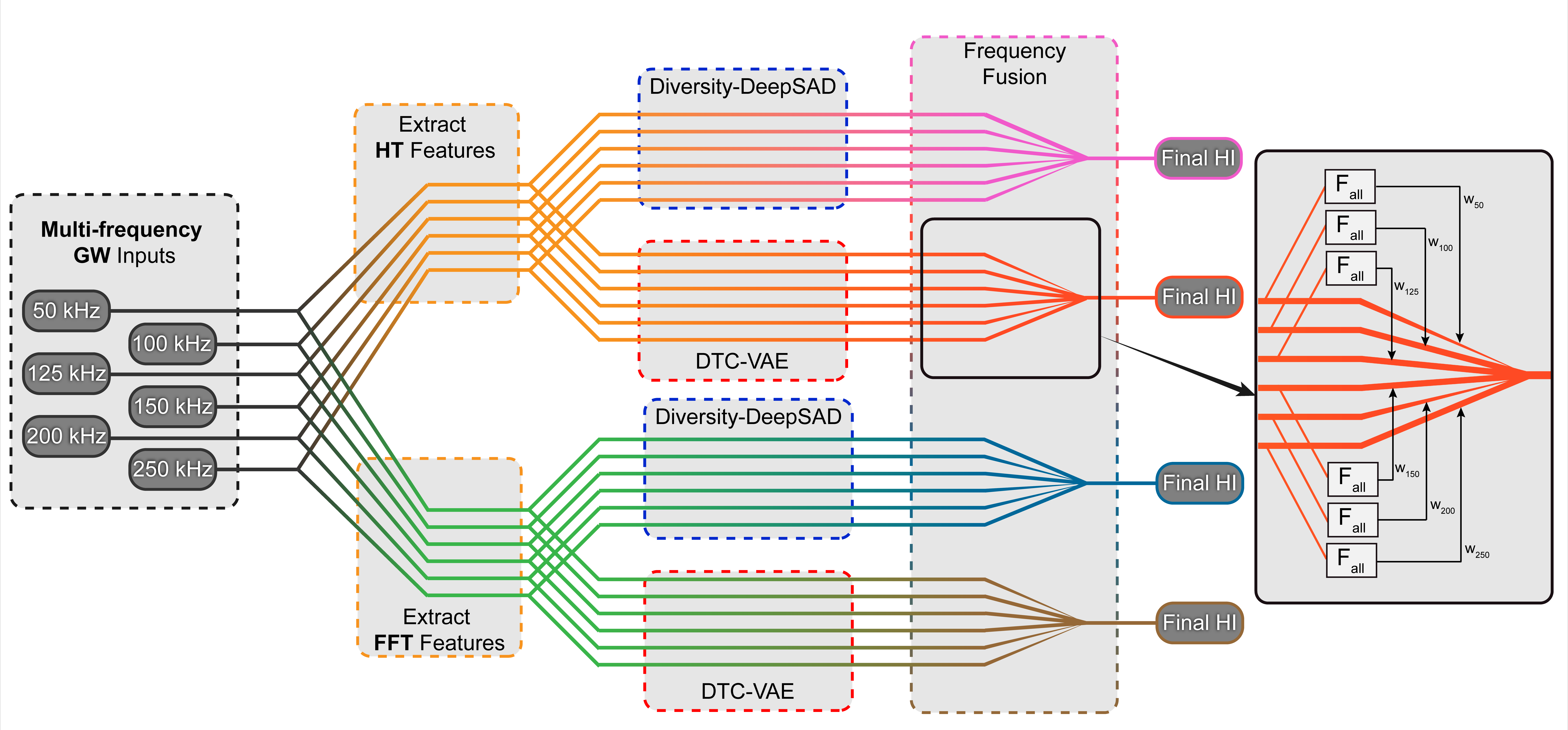}
    \caption{\textcolor{black}{Schematic of the frequency-fusion pipeline, where multi-frequency guided waves are transformed into FFT- and HT-based features to generate per-frequency HIs, which are fused using prognostic-criteria–optimised weights.}}
    \label{fig:FreqFuse}
\end{figure}

\noindent \textcolor{black}{\noindent To summarise the framework, \hyperref[alg:HI_pipeline]{Algorithm 1} presents the full HI generation workflow. It links SP and feature selection with per-frequency model training under leave-one-unit-out cross-validation, prognostic metric computation across random initialisation seeds, and the final weighted frequency-fusion stage. The information pipeline during this process is also shown visually in \autoref{fig:FreqFuse}. Multi-frequency GW measurements are first processed via feature extraction methods, including the HT and FFT, to obtain informative signal representations. These features are then used as input to the proposed models, augmented Diversity-DeepSAD and DTC-VAE, to generate per-frequency HIs. Frequency fusion, performed using WAE, integrates the per-frequency HIs into a unified HI for each model-transform combination.}

\begin{algorithm}[!tbh]
\caption{\textcolor{black}{Overview of HI generation and frequency-fusion framework}} \label{alg:HI_pipeline}
\DontPrintSemicolon \KwIn{\textcolor{black}{Guided wave measurements for all specimens and excitation frequencies}} \KwOut{\textcolor{black}{Per-frequency and fused HI trajectories with associated prognostic metrics}} \BlankLine \textbf{\textcolor{black}{Stage 2-3: Feature extraction and selection}}\; \ForEach{\textcolor{black}{GW measurement (specimen, frequency, path, timestep)}}{ \textcolor{black}{apply SP methods (FFT, STFT, EMD, HT)\; extract statistical features\;} } \ForEach{\textcolor{black}{candidate feature}}{\textcolor{black}{aggregate per timestep and specimen; compute $Mo$, $Pr$, $Tr$ and $F_{all}$ for feature\;}} \textcolor{black}{retain features with fitness above the mean\; select the best-performing SP methods (i.e.\ FFT, HT)\;} \BlankLine \textbf{\textcolor{black}{Stage 4a: Per-frequency HI generation}}\; \ForEach{\textcolor{black}{selected SP method}}{ \ForEach{\textcolor{black}{GW excitation frequency $f$}}{ \ForEach{\textcolor{black}{LOOCV fold (choose one specimen as test, remaining specimens as train)}}{ \textcolor{black}{build training and test feature sets for all timesteps\;} \ForEach{\textcolor{black}{model $\in$ \{Diversity-DeepSAD, DTC-VAE\}}}{ \textcolor{black}{fine-tune hyperparameters by Bayesian optimisation, maximising $F_{all}$ given only training specimens\;} \ForEach{\textcolor{black}{random seed number}}{ \textcolor{black}{train the fine-tuned model using the random seed number and generate per-frequency HIs $y^{m,f}_{(\text{seed})}(t)$ for all specimens\;} \textcolor{black}{compute $Mo$, $Pr$, $Tr$, $F_{all}$ and $Mo_{test}$, $Pr_{test}$, $F_{test}$ for this combination (model, SP, $f$, fold, seed) and store the non-fused HIs and metrics\;} } \textcolor{black}{aggregate prognostic metrics across seeds (mean and standard deviation) for this combination (model, SP, $f$, fold) for reporting\;} } } } } \BlankLine \textbf{\textcolor{black}{Stage 4b: Frequency fusion (WAE)}}\; \ForEach{\textcolor{black}{selected SP method}}{ \ForEach{\textcolor{black}{model $\in$ \{Diversity-DeepSAD, DTC-VAE\}}}{ \ForEach{\textcolor{black}{LOOCV fold}}{ \ForEach{\textcolor{black}{random seed number}}{ \textcolor{black}{apply WAE to per-frequency HIs $y^{m,f}_{(\text{seed})}(t)$ across all GW excitation frequencies $f$ to obtain fused HIs $\bar{y}^{m}_{(\text{seed})}(t)$ for all specimens\;} \textcolor{black}{compute $Mo$, $Pr$, $Tr$, $F_{all}$ and $Mo_{test}$, $Pr_{test}$, $F_{test}$ for this fused combination (model, SP, fold, seed) and store fused HIs and metrics\;} } \textcolor{black}{aggregate fused prognostic metrics across seeds (mean and standard deviation) for this combination (model, SP, fold) \;} } } } \Return{\textcolor{black}{Per-frequency and fused HIs together with their prognostic metrics}}\; \end{algorithm}

\subsection{Hyperparameter sensitivity analysis} \label{subsec:sensitivity_method} \noindent

\noindent \textcolor{black}{In addition to Bayesian optimisation, a sensitivity analysis was performed to assess the robustness of Diversity-DeepSAD and DTC-VAE to their hyperparameters, and quantify how perturbations around the nominal hyperparameters affect the resulting HIs.}

\textcolor{black}{For Diversity-DeepSAD, the analysis focused on the three weighting hyperparameters in \autoref{eq:l_ds} and \autoref{eqn:deepsadloss}: the $L_2$ regularisation weight $\nu$, the auxiliary label weighting $\eta$, and the diversity term weight $\lambda$. Other hyperparameters were fixed to their per-frequency optimal values as baseline. A three-dimensional grid search was then performed over $\nu$, $\eta$ and $\lambda$ on a logarithmic scale, $\{10^{-3}, 10^{-2}, 10^{-1}, 1, 10\}$ for each parameter. For every triplet $(\nu,\eta,\lambda)$ and each excitation frequency, the model was retrained using the same cross-validation folds as in the main experiments, and the HIs were recomputed for all specimens.}
\textcolor{black}{The prognostic criteria in Section~\ref{ss:pc} were then evaluated to obtain both $F_{all}$ and $F_{test}$, and the scores were aggregated by averaging across folds and excitation frequencies. This allowed response surfaces $F_{all}(\eta,\nu|\lambda)$ and $F_{test}(\eta,\nu|\lambda)$ to be constructed for each value of $\lambda$, allowing the identification of regions of stable high performance and the verification of the configuration used in the main study.}

\textcolor{black}{As for DTC-VAE, a similar procedure was followed for the loss weights $\alpha$, $\beta$ and $\gamma$ in \autoref{eq:vae_loss_function}. Similarly, other hyperparameters were fixed to optimal parameters found during Bayesian optimisation to create a per-frequency baseline, allowing a comparison where only the loss weights were altered.}
\textcolor{black}{A regular grid was then sampled over $(\alpha,\beta,\gamma)$ beyond the ranges reported in \autoref{tab:hypspace_VAE}. Again, response surfaces $F_{all}(\alpha,\beta|\gamma)$ and $F_{test}(\alpha,\beta|\gamma)$ were plotted for each $\gamma$, demonstrating the effect of each hyperparameter and justifying the convergence and stability in the given hyperparameter space.}
\vspace{-1em}
\section{Results}\label{sec:results}
\noindent In this section, SP methods are first compared based on the HIs criteria calculated from the statistical features extracted for each method. Following this, the results of the semi-supervised Diversity-DeepSAD model and the unsupervised DTC-VAE model are presented and compared using the HIs criteria across all specimens $F_{all}$ (training, validation, and test) as well as for test specimens alone $F_{test}$. Beyond comparing the two deep learning models, the analysis also examines the impact of different excitation GW frequencies, the fusion of all excitation GW frequencies, and the selection of SP methods (focusing on the two top-performing ones).

\subsection{Signal processing}
\begin{figure}[!tbh]
        \centering
        \includegraphics[width=1.0 \linewidth]{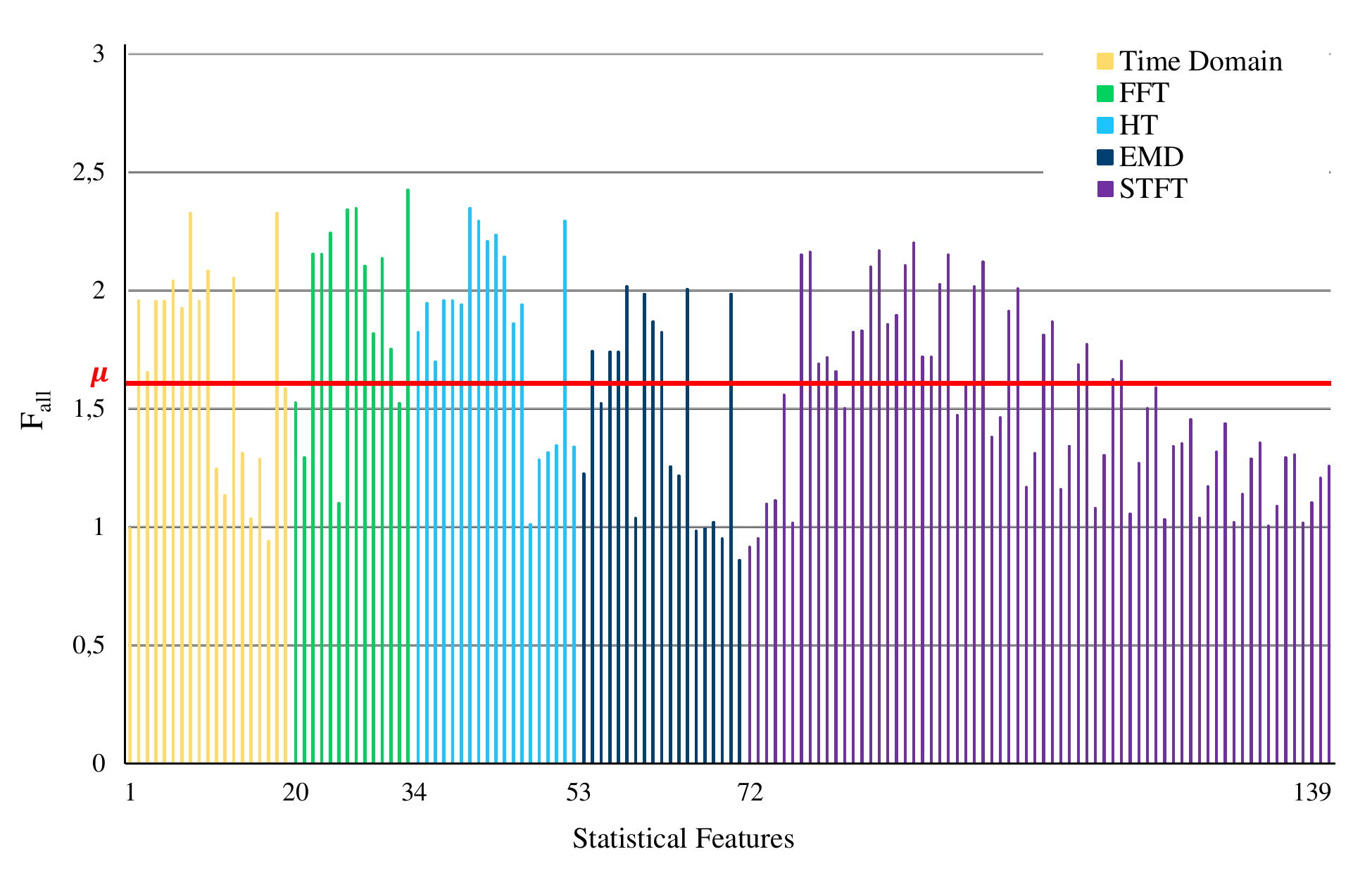}
        \vspace{-20pt}
        \caption{Fitness scores of statistical features extracted from the time domain and other domains derived from various signal processing methods.}
        \label{fig:Exrtacted Feature Scores}
\end{figure}

\noindent \autoref{fig:Exrtacted Feature Scores} shows the $F_{all}$ scores averaged across all paths and frequencies for each of the extracted statistical features. It is clear that all SP methods contain some features that perform poorly when evaluated for fitness. In order to discard the low-performing features, the set is reduced to only those performing above average ($\mu=1.614$), which is indicated by the red horizontal line in \autoref{fig:Exrtacted Feature Scores}. The mean score of features extracted from each SP method is presented in \autoref{tab:MeanOfMethod}. It can be seen that after the feature reduction, FFT and HT yield the highest-performing features, which are used in the subsequent AI models.

\begin{table}[H]
\centering
\caption{Mean fitness scores for each signal processing method, calculated before and after feature reduction.}
\vspace{-5pt}
\resizebox{\textwidth}{!}{%
\begin{tabular}{@{}cccccc@{}}
\toprule
\textbf{} & \textbf{Time Domain}     & \textbf{FFT}     & \textbf{HT}     & \textbf{EMD}     & \textbf{STFT}     \\ \midrule

\textbf{Mean $F_{all}$ before reduction}                 & 1.673 (± 0.45)       & 1.924 (± 0.41)       & 1.840 (± 0.39)       & 1.474 (± 0.42)       & 1.511 (± 0.37)       \\\midrule
\textbf{Mean $F_{all}$ after reduction}             & 2.022 (± 0.18)       & \textbf{2.149} (± 0.21)       & \textbf{2.046} (± 0.20)       & 1.880 (± 0.11)       & 1.910 (± 0.19)    \\ \bottomrule
\end{tabular}%
}
\label{tab:MeanOfMethod}
\end{table}

\subsection{Diversity-DeepSAD}\label{ss:Diversity-DeepSADresults}

\begin{figure}[!tbh]
    \centering
    \includegraphics[width=\textwidth]{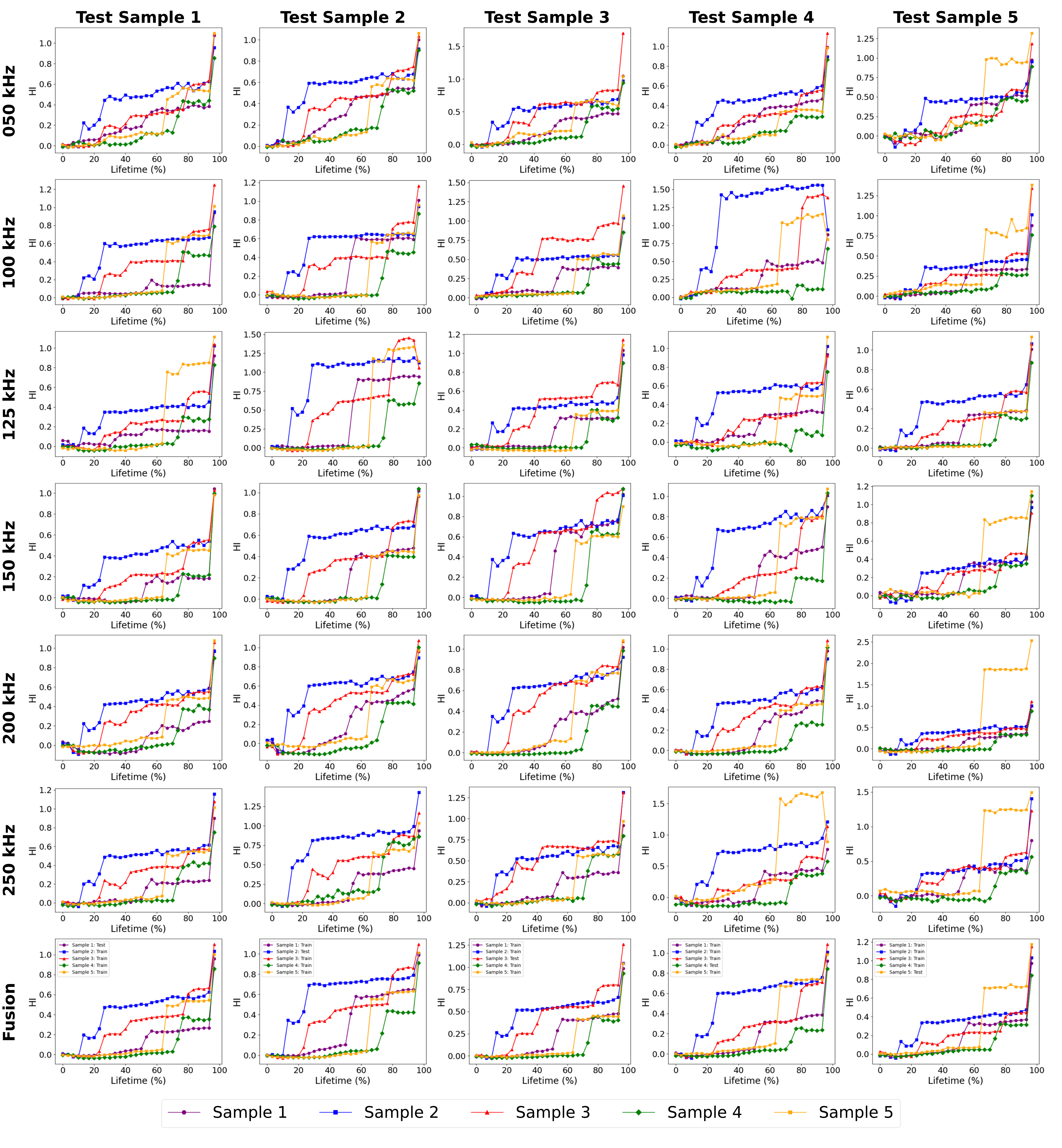}
    \vspace{-20pt}
    \caption{HIs constructed by Diversity-DeepSAD using FFT features.}
    \label{fig:DeepSAD_FFT}
\end{figure}

\begin{figure}[!tbh]
    \centering
    \includegraphics[width=\textwidth]{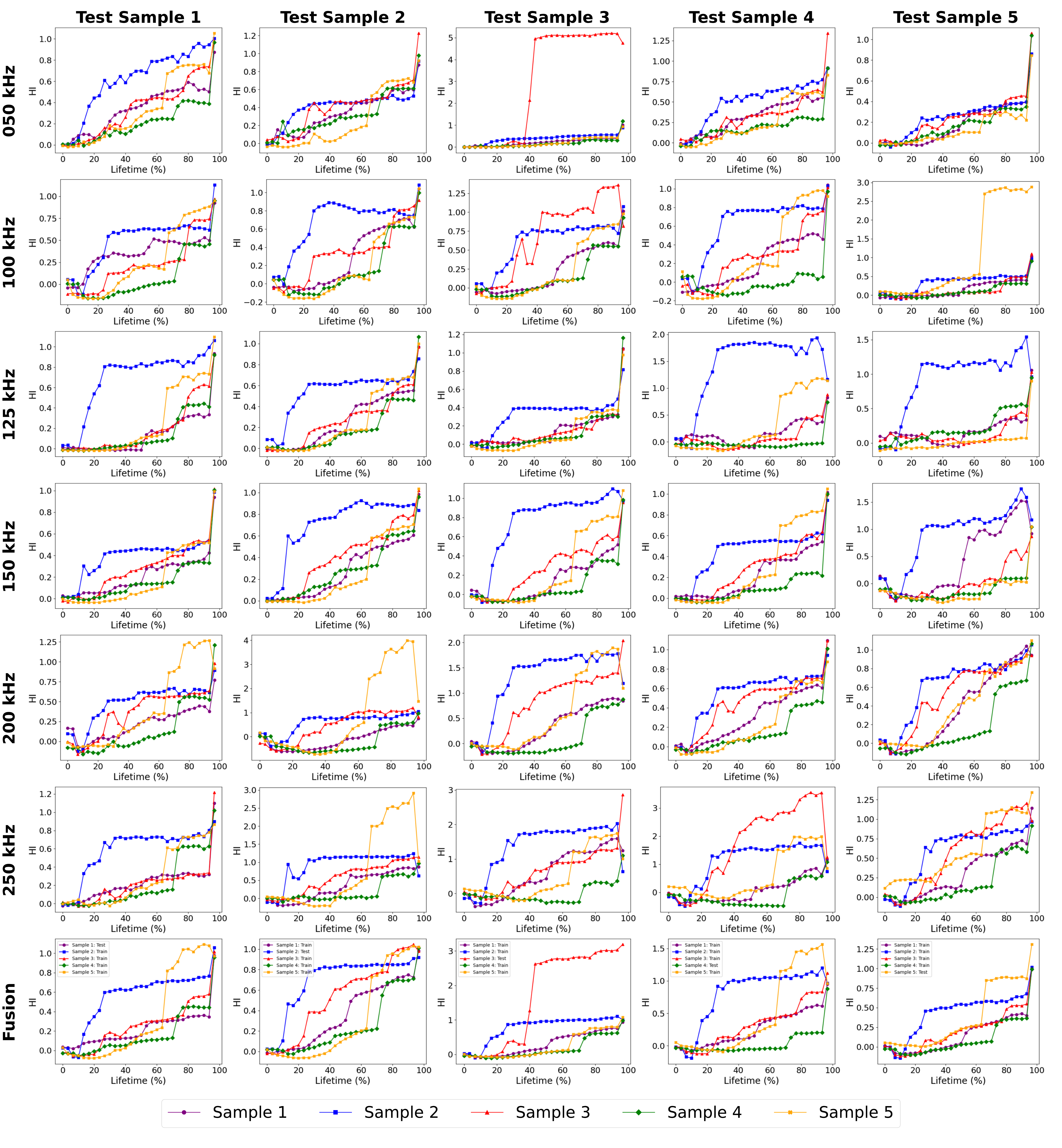}
    \vspace{-20pt}
    \caption{HIs constructed by Diversity-DeepSAD using HT features.}
    \label{fig:DeepSAD_HLB}
\end{figure}

\noindent The Diversity\textcolor{black}{-}DeepSAD model was trained across all combinations of frequencies and data folds (i.e.\ training and test divisions), with WAE implemented as described. 
The HIs constructed by Diversity\textcolor{black}{-}DeepSAD upon FFT and HT features are shown in \autoref{fig:DeepSAD_FFT} and \autoref{fig:DeepSAD_HLB}, respectively. For better comparison, the HIs are displayed over the normalised lifetime of specimens, ranging from 0\% to 100\%.   
The resulting fitness scores of the HIs are reported in \autoref{tab:Comparison_DeepSAD_FFT_HLB_F-all} and \autoref{tab:Comparison_DeepSAD_FFT_HT_F-test}. These tables include $F_{all}$ scores, from \autoref{eq:fitness}, and $F_{test}$ scores, from \autoref{eq:ftest}, for HIs generated from both FFT and HT features.

Diversity-DeepSAD results show high fitness scores for both FFT and HT, with mean $F_{test}$ scores of 2.27 and 2.13 respectively, with $F_{all}$ scores averaging slightly higher at 2.35 and 2.22. This implies a general higher performance of features from the FFT than those from the HT. No fold performed consistently \textcolor{black}{worse} than the others, although average results from test Sample 1 were consistently highest. Inspecting the graph, this may be due to the invariably conforming behaviour of this sample, aiding in high $F_{test}$ scores. All frequencies performed similarly, with \SI{50}{\kilo\hertz} producing marginally better average $F_{all}$ results for both SP methods.

While the range of scores is similar for both sets of input data, FFT demonstrates a greater robustness due to its low standard deviation between random seeds, with an $F_{test}$ mean $\bar{\sigma}=$0.10 compared to 0.14 for HT. This indicates a stronger robustness in this model using the FFT features than using those of the HT.

\begin{table}[!bh]
\centering
\caption{Fitness scores across all units ($F_{all}$) for the Diversity-DeepSAD model using FFT and HT features over 5 iterations.}
\vspace{-5pt}
\resizebox{\textwidth}{!}{%
\begin{tabular}{@{}ccccccccccc@{}}
\toprule
\textbf{f {[}kHz{]}} & \multicolumn{2}{c}{\textbf{Fold 1}} & \multicolumn{2}{c}{\textbf{Fold 2}} & \multicolumn{2}{c}{\textbf{Fold 3}} & \multicolumn{2}{c}{\textbf{Fold 4}} & \multicolumn{2}{c}{\textbf{Fold 5}} \\ 
\cmidrule(lr){2-3} \cmidrule(lr){4-5} \cmidrule(lr){6-7} \cmidrule(lr){8-9} \cmidrule(lr){10-11}
                            & \textbf{FFT}   & \textbf{HT}    & \textbf{FFT}   & \textbf{HT}    & \textbf{FFT}   & \textbf{HT}    & \textbf{FFT}   & \textbf{HT}    & \textbf{FFT}   & \textbf{HT}    \\ \midrule
\textbf{50} & 
2.55 (± 0.04) & 2.29 (± 0.28) &
2.47 (± 0.03) & 2.49 (± 0.03) &
2.40 (± 0.05) & 1.85 (± 0.06) &
2.50 (± 0.06) & 2.47 (± 0.07) &
2.47 (± 0.06) & 2.67 (± 0.08) \\
\textbf{100} &
2.34 (± 0.07) & 2.30 (± 0.05) &
2.22 (± 0.05) & 2.19 (± 0.14) &
2.36 (± 0.04) & 2.31 (± 0.04) &
1.92 (± 0.19) & 2.22 (± 0.10) &
2.47 (± 0.03) & 1.99 (± 0.11) \\
\textbf{125} &
2.43 (± 0.06) & 2.02 (± 0.22) &
2.20 (± 0.12) & 2.42 (± 0.04) &
2.38 (± 0.03) & 2.37 (± 0.08) &
2.42 (± 0.10) & 2.07 (± 0.23) &
2.47 (± 0.05) & 2.18 (± 0.33) \\
\textbf{150} &
2.38 (± 0.12) & 2.56 (± 0.11) &
2.41 (± 0.04) & 2.42 (± 0.06) &
2.16 (± 0.14) & 2.28 (± 0.08) &
2.06 (± 0.11) & 2.37 (± 0.14) &
2.49 (± 0.06) & 2.06 (± 0.14) \\
\textbf{200} &
2.39 (± 0.05) & 2.35 (± 0.04) &
2.32 (± 0.03) & 2.04 (± 0.17) &
2.34 (± 0.06) & 1.82 (± 0.09) &
2.51 (± 0.05) & 2.33 (± 0.13) &
2.32 (± 0.11) & 2.51 (± 0.03) \\
\textbf{250} &
2.36 (± 0.05) & 2.26 (± 0.04) &
2.30 (± 0.05) & 1.95 (± 0.13) &
2.32 (± 0.05) & 1.92 (± 0.32) &
2.14 (± 0.07) & 1.67 (± 0.19) &
2.28 (± 0.10) & 2.25 (± 0.15) \\
\midrule
\textbf{Fusion} & 
2.50 (± 0.03) & 2.51 (± 0.02) &
2.42 (± 0.03) & 2.37 (± 0.07) &
2.44 (± 0.02) & 2.01 (± 0.08) &
2.38 (± 0.05) & 2.22 (± 0.12) &
2.51 (± 0.05) & 2.44 (± 0.06) \\
\bottomrule
\end{tabular}%
}
\label{tab:Comparison_DeepSAD_FFT_HLB_F-all}
\end{table}

\begin{table}[!tbh]
\centering
\caption{Test fitness scores ($F_{test}$) for the Diversity-DeepSAD model using FFT and HT features over 5 iterations.}
\vspace{-5pt}
\resizebox{\textwidth}{!}{%
\begin{tabular}{@{}ccccccccccc@{}}
\toprule
\textbf{f {[}kHz{]}} & \multicolumn{2}{c}{\textbf{Fold 1}} & \multicolumn{2}{c}{\textbf{Fold 2}} & \multicolumn{2}{c}{\textbf{Fold 3}} & \multicolumn{2}{c}{\textbf{Fold 4}} & \multicolumn{2}{c}{\textbf{Fold 5}} \\ 
\cmidrule(lr){2-3} \cmidrule(lr){4-5} \cmidrule(lr){6-7} \cmidrule(lr){8-9} \cmidrule(lr){10-11}
                            & \textbf{FFT}   & \textbf{HT}    & \textbf{FFT}   & \textbf{HT}    & \textbf{FFT}   & \textbf{HT}    & \textbf{FFT}   & \textbf{HT}    & \textbf{FFT}   & \textbf{HT}    \\ \midrule
\textbf{50}  &
2.63 (± 0.06) & 2.25 (± 0.24) &
2.44 (± 0.04) & 2.48 (± 0.02) &
2.30 (± 0.08) & 1.57 (± 0.09) &
2.41 (± 0.08) & 2.48 (± 0.07) &
2.39 (± 0.13) & 2.53 (± 0.10) \\
\textbf{100} &
2.37 (± 0.08) & 2.26 (± 0.05) &
2.17 (± 0.12) & 2.00 (± 0.15) &
2.21 (± 0.06) & 2.12 (± 0.06) &
1.67 (± 0.22) & 2.13 (± 0.11) &
2.40 (± 0.08) & 1.64 (± 0.10) \\
\textbf{125} &
2.42 (± 0.12) & 1.99 (± 0.18) &
2.14 (± 0.17) & 2.28 (± 0.08) &
2.35 (± 0.07) & 2.43 (± 0.11) &
2.30 (± 0.15) & 2.03 (± 0.30) &
2.36 (± 0.05) & 2.21 (± 0.27) \\
\textbf{150} &
2.29 (± 0.11) & 2.58 (± 0.14) &
2.42 (± 0.04) & 2.27 (± 0.10) &
2.31 (± 0.10) & 2.32 (± 0.06) &
1.75 (± 0.08) & 2.30 (± 0.17) &
2.35 (± 0.07) & 1.93 (± 0.12) \\
\textbf{200} &
2.40 (± 0.05) & 2.31 (± 0.08) &
2.28 (± 0.06) & 2.02 (± 0.23) &
2.31 (± 0.05) & 1.63 (± 0.11) &
2.46 (± 0.04) & 2.38 (± 0.15) &
2.07 (± 0.25) & 2.47 (± 0.03) \\
\textbf{250} &
2.36 (± 0.08) & 2.18 (± 0.11) &
2.17 (± 0.06) & 1.59 (± 0.15) &
2.23 (± 0.13) & 1.77 (± 0.44) &
1.88 (± 0.11) & 1.56 (± 0.13) &
2.17 (± 0.12) & 2.15 (± 0.20) \\
\midrule
\textbf{Fusion} &
2.56 (± 0.03) & 2.50 (± 0.03) &
2.44 (± 0.04) & 2.31 (± 0.07) &
2.35 (± 0.05) & 1.74 (± 0.06) &
2.25 (± 0.08) & 2.20 (± 0.09) &
2.37 (± 0.09) & 2.18 (± 0.09) \\
\bottomrule
\end{tabular}%
}
\label{tab:Comparison_DeepSAD_FFT_HT_F-test}
\end{table}

\subsection{DTC-VAE}\label{ss:VAE results}
\begin{figure}[!tbh]
    \centering
    \includegraphics[width=\textwidth]{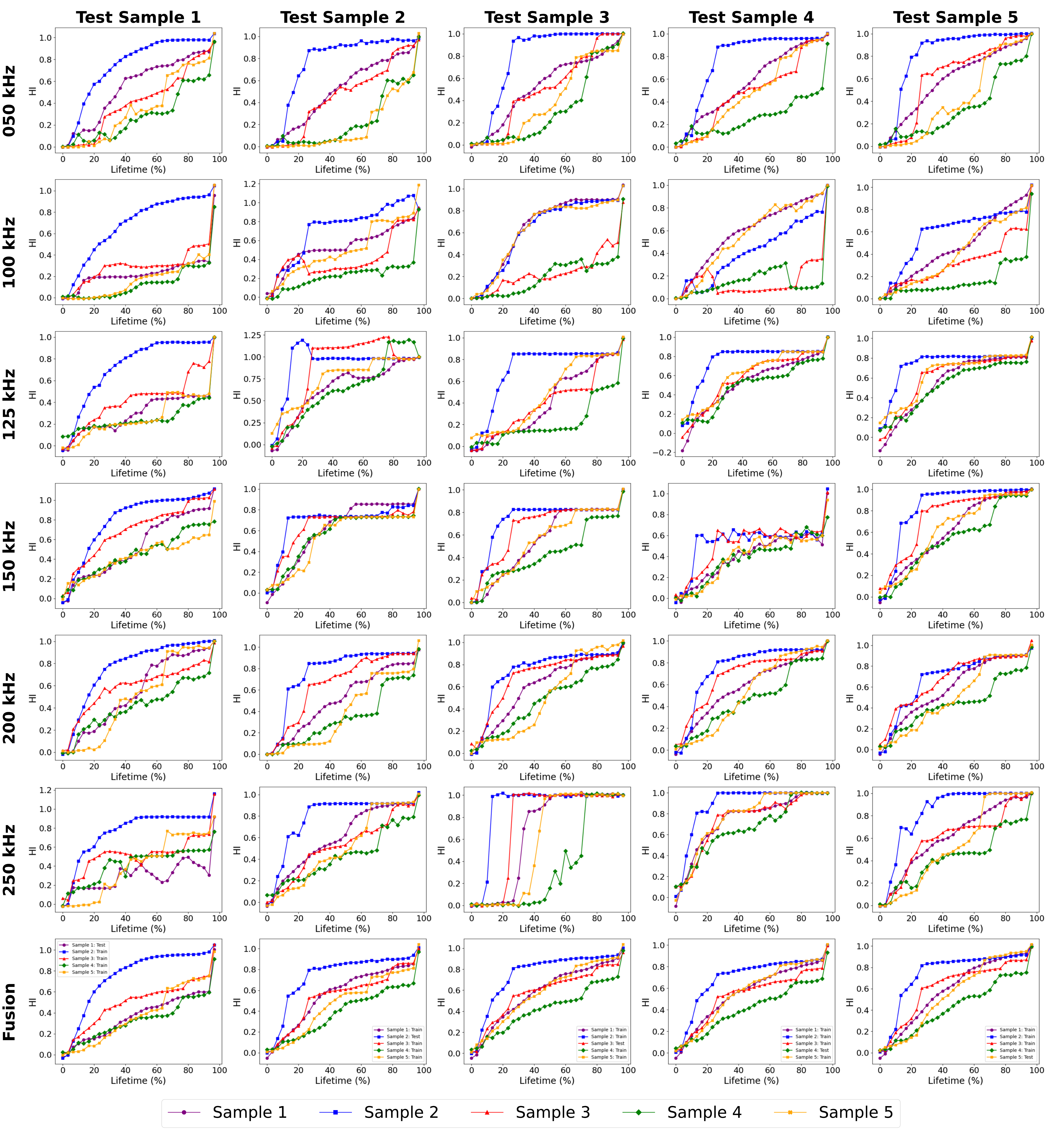}
    \vspace{-20pt}
    \caption{HIs constructed by DTC-VAE using FFT features.}
    \label{fig:VAE_FFT}
\end{figure}

\begin{figure}[!tbh]
    \centering
    \includegraphics[width=\textwidth]{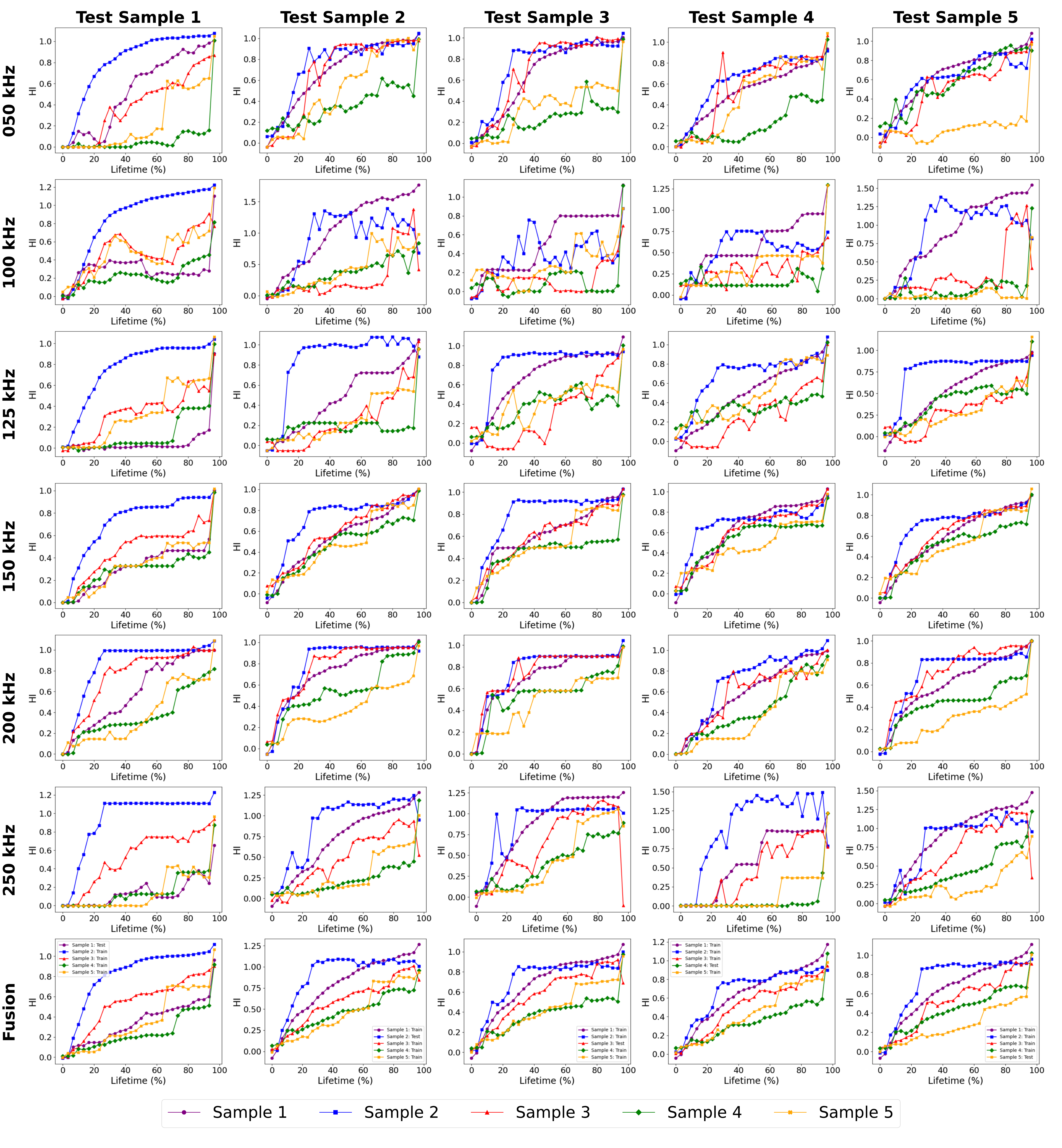}
    \vspace{-20pt}
    \caption{HIs constructed by DTC-VAE using HT features.}
    \label{fig:VAE_hlb}
\end{figure}

\noindent The DTC-VAE model was similarly trained and tested on all combinations of frequencies and folds for 5 random seeds. 
The HIs extracted by DTC-VAE from both FFT and HT features are displayed in \autoref{fig:VAE_FFT} and \autoref{fig:VAE_hlb}, respectively.
The resulting fitness scores of the HIs generated by the DTC-VAE are thus provided in \autoref{tab:Comparison_FFT_HT_F-all} and \ref{tab:Comparison_FFT_HT_F-test}. These tables cover both FFT and HT features, detailing scores for the general fitness $F_{all}$ and test fitness $F_{test}$. 

FFT features achieved consistently high results, with a mean $F_{test}$ scores of 2.55 and 2.20 for the FFT and HT respectively, and similar $F_{all}$ scores of 2.56 and 2.25. These imply features from the FFT generate again generally better performing HIs than the HT. No fold produced consistently different scores to the others; inspection of the graphs reveals distributed anomalies for all. However, for DTC-VAE, HIs generated using features from the \SI{100}{\kilo\hertz} signal tended to perform poorly, with an average $F_{test}$ score of 2.06 compared to the overall mean score of 2.38.

FFT features exhibited higher stability than those of the HT, with an overall lowest $F_{test}$ score of 2.24 compared to 1.42. This is further highlighted by the low standard deviation values, with an $F_{test}$ mean $\bar{\sigma}=$0.09 compared to 0.12 for HT.
The stronger robustness in both models using the FFT features than those of the HT indicates that the former include more reliable information on the structure's health, consistent with the results in \autoref{tab:MeanOfMethod}.

\begin{table}[!tbh]
\centering
\caption{Fitness scores across all units ($F_{all}$) for the DTC-VAE model using FFT and HT features over 5 iterations.}
\vspace{-5pt}
\resizebox{\textwidth}{!}{%
\begin{tabular}{@{}ccccccccccc@{}}
\toprule
\textbf{f [kHz]} & \multicolumn{2}{c}{\textbf{Fold 1}} & \multicolumn{2}{c}{\textbf{Fold 2}} & \multicolumn{2}{c}{\textbf{Fold 3}} & \multicolumn{2}{c}{\textbf{Fold 4}} & \multicolumn{2}{c}{\textbf{Fold 5}} \\ 
\cmidrule(lr){2-3}\cmidrule(lr){4-5}\cmidrule(lr){6-7}\cmidrule(lr){8-9}\cmidrule(lr){10-11}
 & \textbf{FFT} & \textbf{HT} & \textbf{FFT} & \textbf{HT} & \textbf{FFT} & \textbf{HT} & \textbf{FFT} & \textbf{HT} & \textbf{FFT} & \textbf{HT} \\ 
\midrule
\textbf{50} & 
2.66 (± 0.03) & 2.16 (± 0.11) & 
2.45 (± 0.05) & 2.55 (± 0.07) & 
2.57 (± 0.04) & 2.42 (± 0.06) & 
2.52 (± 0.11) & 2.40 (± 0.06) & 
2.50 (± 0.05) & 2.09 (± 0.14) \\

\textbf{100} & 
2.37 (± 0.11) & 1.96 (± 0.11) &
2.41 (± 0.19) & 1.88 (± 0.13) &
2.55 (± 0.09) & 1.71 (± 0.19) &
2.52 (± 0.14) & 1.61 (± 0.20) &
2.45 (± 0.08) & 1.50 (± 0.15) \\

\textbf{125} & 
2.43 (± 0.07) & 2.00 (± 0.13) &
2.49 (± 0.11) & 2.11 (± 0.11) &
2.54 (± 0.07) & 2.20 (± 0.13) &
2.57 (± 0.13) & 2.30 (± 0.10) &
2.57 (± 0.05) & 2.27 (± 0.04) \\

\textbf{150} & 
2.69 (± 0.05) & 2.63 (± 0.10) &
2.54 (± 0.09) & 2.59 (± 0.05) &
2.53 (± 0.11) & 2.60 (± 0.07) &
2.51 (± 0.08) & 2.62 (± 0.03) &
2.75 (± 0.02) & 2.65 (± 0.05) \\

\textbf{200} & 
2.81 (± 0.03) & 2.50 (± 0.07) &
2.55 (± 0.10) & 2.51 (± 0.02) &
2.73 (± 0.06) & 2.62 (± 0.08) &
2.77 (± 0.03) & 2.64 (± 0.07) &
2.79 (± 0.04) & 2.52 (± 0.08) \\

\textbf{250} & 
2.35 (± 0.11) & 2.06 (± 0.06) &
2.66 (± 0.05) & 2.19 (± 0.07) &
2.44 (± 0.21) & 1.97 (± 0.04) &
2.51 (± 0.13) & 1.99 (± 0.25) &
2.67 (± 0.06) & 2.16 (± 0.05) \\
\midrule
\textbf{Fusion} &
2.75 (± 0.01) & 2.55 (± 0.02) &
2.72 (± 0.05) & 2.51 (± 0.07) &
2.77 (± 0.07) & 2.56 (± 0.05) &
2.80 (± 0.03) & 2.61 (± 0.04) &
2.80 (± 0.03) & 2.49 (± 0.02) \\
\bottomrule
\end{tabular}%
}
\label{tab:Comparison_FFT_HT_F-all}
\end{table}

\begin{table}[!tbh]
\centering
\caption{Test fitness scores ($F_{test}$) for the DTC-VAE model using FFT and HT features over 5 iterations.}
\vspace{-5pt}
\resizebox{\textwidth}{!}{%
\begin{tabular}{@{}ccccccccccc@{}}
\toprule
\textbf{f [kHz]} & \multicolumn{2}{c}{\textbf{Fold 1}} & \multicolumn{2}{c}{\textbf{Fold 2}} & \multicolumn{2}{c}{\textbf{Fold 3}} & \multicolumn{2}{c}{\textbf{Fold 4}} & \multicolumn{2}{c}{\textbf{Fold 5}} \\ 
\cmidrule(lr){2-3}\cmidrule(lr){4-5}\cmidrule(lr){6-7}\cmidrule(lr){8-9}\cmidrule(lr){10-11}
 & \textbf{FFT} & \textbf{HT} & \textbf{FFT} & \textbf{HT} & \textbf{FFT} & \textbf{HT} & \textbf{FFT} & \textbf{HT} & \textbf{FFT} & \textbf{HT} \\ 
\midrule

\textbf{50}  &
2.68 (± 0.02) & 2.20 (± 0.13) &
2.41 (± 0.07) & 2.50 (± 0.08) &
2.57 (± 0.07) & 2.35 (± 0.10) &
2.52 (± 0.08) & 2.39 (± 0.08) &
2.53 (± 0.04) & 2.03 (± 0.15) \\

\textbf{100} &
2.34 (± 0.21) & 1.92 (± 0.08) &
2.36 (± 0.15) & 1.98 (± 0.11) &
2.49 (± 0.13) & 1.42 (± 0.30) &
2.41 (± 0.21) & 1.54 (± 0.19) &
2.52 (± 0.07) & 1.61 (± 0.21) \\

\textbf{125} &
2.43 (± 0.06) & 1.86 (± 0.19) &
2.43 (± 0.14) & 2.00 (± 0.13) &
2.58 (± 0.08) & 2.32 (± 0.08) &
2.65 (± 0.10) & 2.22 (± 0.14) &
2.62 (± 0.08) & 2.27 (± 0.06) \\

\textbf{150} &
2.72 (± 0.05) & 2.66 (± 0.12) &
2.35 (± 0.16) & 2.44 (± 0.12) &
2.58 (± 0.11) & 2.65 (± 0.05) &
2.51 (± 0.08) & 2.53 (± 0.05) &
2.75 (± 0.04) & 2.62 (± 0.06) \\

\textbf{200} &
2.83 (± 0.02) & 2.58 (± 0.09) &
2.51 (± 0.15) & 2.26 (± 0.10) &
2.73 (± 0.06) & 2.57 (± 0.07) &
2.78 (± 0.02) & 2.65 (± 0.07) &
2.76 (± 0.05) & 2.61 (± 0.05) \\

\textbf{250} &
2.24 (± 0.07) & 1.90 (± 0.11) &
2.60 (± 0.11) & 2.13 (± 0.08) &
2.44 (± 0.27) & 1.55 (± 0.06) &
2.58 (± 0.12) & 2.05 (± 0.33) &
2.65 (± 0.05) & 2.35 (± 0.09) \\

\midrule

\textbf{Fusion} &
2.76 (± 0.02) & 2.60 (± 0.03) &
2.71 (± 0.05) & 2.43 (± 0.09) &
2.76 (± 0.08) & 2.41 (± 0.08) &
2.77 (± 0.02) & 2.61 (± 0.04) &
2.81 (± 0.03) & 2.61 (± 0.03) \\

\bottomrule
\end{tabular}%
}
\label{tab:Comparison_FFT_HT_F-test}
\end{table}

\subsection{Weighted Average Ensemble}

\noindent Overall, the fused models show improved results compared to single frequencies, with generally higher scores consistently outperforming the corresponding average fitness score of a particular fold. Across both FFT and HT, the average unfused $F_{all}$ and $F_{test}$ scores for Diversity-DeepSAD are 2.29 and 2.20, while for DTC-VAE they are 2.41 and 2.38 respectively. The same fused scores average 2.38, 2.29, 2.66 and 2.65. This is an average score increase of 3.2\% for Diversity-DeepSAD, but 10.9\% for DTC-VAE, with both $F_{all}$ and $F_{test}$ scores showing similar improvement. The high performance of WAE, fusing different GW excitation frequencies, combined with DTC-VAE is also evident from the graphs, where the fused HIs consistently behave more cleanly than their constituent frequencies.

The stability of fused results is also greatly improved, with deviations between folds smaller than within individual frequencies. For both models, but still most noticeably DTC-VAE, the fused graphs are visually more consistent between folds than unfused. Between seeds, the standard deviations of fitness for the fusion are always below 0.1, with one exception in Diversity-DeepSAD fold 4. The average standard deviation for Diversity-DeepSAD-Fusion scores is half that of unfused, 0.05 compared to 0.10 for $F_{all}$ and 0.06 to 0.12 for $F_{test}$. For DTC-VAE, the reduction is better than half at 0.04 from 0.09 and 0.05 from 0.11 respectively. These low standard deviations indicate that HIs fused using WAE of multiple GW excitation frequencies are more robust. This is because different GW excitation frequencies can capture various damage modes in structures, and the ensemble approach reduces susceptibility to errors in individual constituent models.

Across all results, there are only three occurrences where the standard deviation of fused results for HIs from the FFT exceeds that of the HT, confirming that the features extracted from the output of FFT also tend to produce more consistently performing HIs. The $F_{test}$ scores for Diversity-DeepSAD with WAE average 2.39 and 2.19 for FFT and HT respectively, while DTC-VAE shows a large increase to 2.76 and 2.53. Every DTC-VAE WAE fitness score is higher than its Diversity-DeepSAD counterpart, showing exceptionally high performance at a peak of 2.81 or 94\%.

\subsection{Model comparison} \label{sec:model_comparison}
\noindent The prognostic criteria and fitness scores (reported in the previous subsections and tables) were averaged across all folds for both models to comprehensively perform a comparison, including the WAE models. The resulting averages for $F_{all}$, along with the prognostic criteria scores, are displayed in \autoref{tab:f_all_average}, while those for $F_{test}$ are included in \autoref{tab:f_test_average}.

\begin{table}[!tbh]
\centering
\caption{Prognostic criteria and fitness scores ($F_{all}$) averaged across all folds.}
\vspace{-5pt}
\resizebox{\textwidth}{!}{
\begin{tabular}{@{}cccccccccc@{}}
\toprule
\textbf{Score} & \multicolumn{2}{c}{\textbf{Diversity-DeepSAD}} & \multicolumn{2}{c}{\textbf{Diversity-DeepSAD WAE}} & \multicolumn{2}{c}{\textbf{DTC-VAE}} & \multicolumn{2}{c}{\textbf{DTC-VAE WAE}} \\
\cmidrule(lr){2-3} \cmidrule(lr){4-5} \cmidrule(lr){6-7} \cmidrule(lr){8-9}
 & \textbf{FFT} & \textbf{HT} & \textbf{FFT} & \textbf{HT} & \textbf{FFT} & \textbf{HT} & \textbf{FFT} & \textbf{HT} \\
\midrule
$\boldsymbol{Mo}$ & 0.87 (\textpm 0.04) & 0.87 (\textpm 0.06) & 0.94 (\textpm 0.01) & 0.93 (\textpm 0.03) & 0.92 (\textpm 0.06) & 0.85 (\textpm 0.07) & 0.99 (\textpm 0.01) & 0.93 (\textpm 0.03) \\
$\boldsymbol{Pr}$ & 0.89 (\textpm 0.07) & 0.85 (\textpm 0.16) & 0.91 (\textpm 0.03) & 0.84 (\textpm 0.18) & 0.97 (\textpm 0.04) & 0.87 (\textpm 0.13) & 0.98 (\textpm 0.01) & 0.90 (\textpm 0.02) \\
$\boldsymbol{Tr}$ & 0.59 (\textpm 0.13) & 0.50 (\textpm 0.17) & 0.60 (\textpm 0.08) & 0.54 (\textpm 0.11) & 0.68 (\textpm 0.13) & 0.52 (\textpm 0.22) & 0.80 (\textpm 0.05) & 0.71 (\textpm 0.05) \\
\midrule
$\boldsymbol{F_{all}}$ & 2.35 (\textpm 0.16) & 2.22 (\textpm 0.28) & 2.45 (\textpm 0.06) & 2.31 (\textpm 0.20) & 2.56 (\textpm 0.16) & 2.25 (\textpm 0.34) & 2.77 (\textpm 0.05) & 2.54 (\textpm 0.06) \\
\bottomrule
\end{tabular}
}
\label{tab:f_all_average}
\end{table}

\begin{table}[!tbh]
\centering
\caption{Test prognostic criteria and test fitness scores ($F_{test}$) averaged across all folds.}
\vspace{-5pt}
\resizebox{\textwidth}{!}{
\begin{tabular}{@{}cccccccccc@{}}
\toprule
\textbf{Score} & \multicolumn{2}{c}{\textbf{Diversity-DeepSAD}} & \multicolumn{2}{c}{\textbf{Diversity-DeepSAD WAE}} & \multicolumn{2}{c}{\textbf{DTC-VAE}} & \multicolumn{2}{c}{\textbf{DTC-VAE WAE}} \\
\cmidrule(lr){2-3} \cmidrule(lr){4-5} \cmidrule(lr){6-7} \cmidrule(lr){8-9}
 & \textbf{FFT} & \textbf{HT} & \textbf{FFT} & \textbf{HT} & \textbf{FFT} & \textbf{HT} & \textbf{FFT} & \textbf{HT} \\
\midrule
$\boldsymbol{Mo_{test}}$ & 0.83 (\textpm 0.04) & 0.82 (\textpm 0.12) & 0.93 (\textpm 0.04) & 0.91 (\textpm 0.05) & 0.91 (\textpm 0.10) & 0.81 (\textpm 0.13) & 0.98 (\textpm 0.02) & 0.91 (\textpm 0.10) \\
$\boldsymbol{Pr_{test}}$ & 0.85 (\textpm 0.13) & 0.81 (\textpm 0.24) & 0.86 (\textpm 0.10) & 0.74 (\textpm 0.29) & 0.97 (\textpm 0.05) & 0.88 (\textpm 0.16) & 0.97 (\textpm 0.02) & 0.91 (\textpm 0.09) \\
$\boldsymbol{Tr}$ & 0.59 (\textpm 0.13) & 0.50 (\textpm 0.17) & 0.60 (\textpm 0.08) & 0.54 (\textpm 0.11) & 0.68 (\textpm 0.13) & 0.52 (\textpm 0.22) & 0.80 (\textpm 0.05) & 0.71 (\textpm 0.05) \\
\midrule
$\boldsymbol{F_{test}}$ & 2.27 (\textpm 0.23) & 2.13 (\textpm 0.34) & 2.39 (\textpm 0.12) & 2.19 (\textpm 0.26) & 2.55 (\textpm 0.18) & 2.20 (\textpm 0.38) & 2.76 (\textpm 0.06) & 2.53 (\textpm 0.11) \\
\bottomrule
\end{tabular}
}
\label{tab:f_test_average}
\end{table}

Both models showed promising monotonicity scores, with Diversity-DeepSAD achieving an average $Mo$ score of 0.87 and 0.87 for FFT and HT features respectively, while its average $Mo_{test}$ scores were 0.83 and 0.82. DTC-VAE achieved average $Mo$ scores of 0.92 and 0.85 and average $Mo_{test}$ scores 0.91 and 0.81. 

The average prognosability scores were also high, particularly for DTC-VAE. Diversity-DeepSAD achieved an average $Pr$ score of 0.89 and 0.85 for FFT and HT features respectively. Its average $Pr_{test}$ scores were 0.85 and 0.81. On the other hand, DTC-VAE achieved average $Pr$ scores of 0.97 and 0.87, demonstrating exceptional performance on FFT features. Similarly, its average $Pr_{test}$ scores were 0.91 and 0.81.

The results reflect the fact that trendability was the lowest performing prognostic criterion for both models, as can be seen by the inconsistency of shapes between specimens in the HI graphs. Both models achieved moderate scores, with Diversity-DeepSAD scoring an average $Tr$ score of 0.59 and 0.50, and DTC-VAE scoring 0.68 and 0.52, for FFT and HT features respectively. \textcolor{black}{As explained in \autoref{ss: EXP}, each specimen had differences in manufacturing and the presence or absence of manufacturing disbonds or impact events. These factors significantly impact GW propoagation characteristics, and it is therefore physically justified that the $Tr$ score produced was lower than that for $Mo$ or $Pr$.}

Overall, HIs extracted by Diversity-DeepSAD using FFT features performed slightly better than using HT features. Similarly, DTC-VAE also generated better HIs when trained on FFT features, although the difference in performance was greater. However, both models showed significant improvements when combined with WAE, with all prognostic criteria showing improved scores and stability for both FFT and HT features. Diversity-DeepSAD WAE achieved average $F_{all}$ scores of 2.45 and 2.31, while its $F_{test}$ scores were 2.39 and 2.19. In addition, DTC-VAE WAE scored 2.77 and 2.54 in $F_{all}$, and 2.76 and 2.53 in $F_{test}$.
\vspace{-0.7em}
\textcolor{black}{\subsection{Sensitivity analysis of hyperparameters} \label{subsec:sensitivity_results}}

\noindent \textcolor{black}{The hyperparameter sensitivity analysis confirmed that the conclusions drawn in the previous subsections are not the result of fragile or excessively fine-tuned configurations. For both Diversity-DeepSAD and DTC-VAE, the response surfaces constructed for $F_{all}$ and $F_{test}$ show broad regions of consistently high performance around the nominal hyperparameters obtained via Bayesian optimisation.}

\begin{figure}[!tbh]
    \centering

    \begin{subfigure}{0.70\linewidth}
        \centering
        \includegraphics[width=\linewidth]{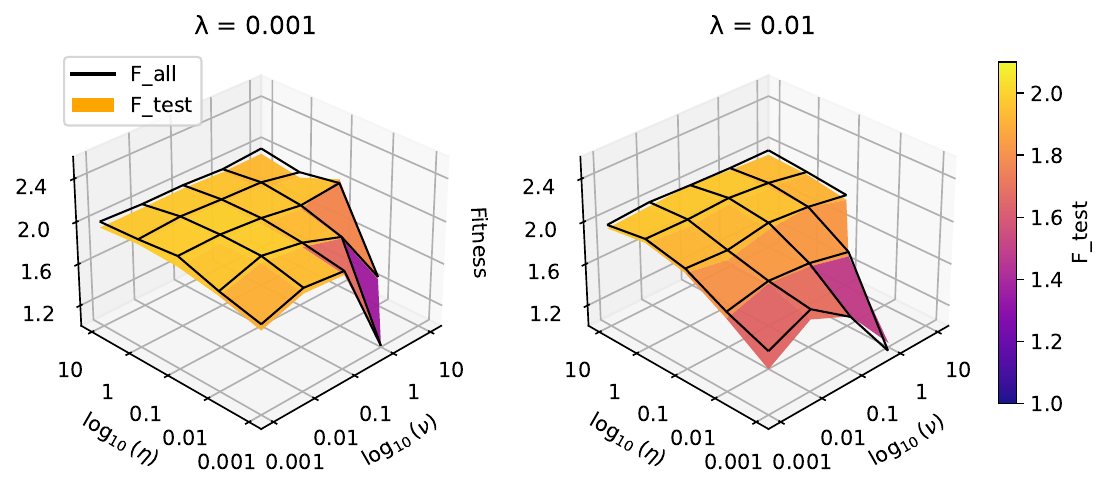}
        \caption{\textcolor{black}{Diversity-DeepSAD using FFT features.}}
        \label{fig:deepsad_fft_sensitivity_reduced}
    \end{subfigure}

    \vspace{8pt}

    \begin{subfigure}{0.70\linewidth}
        \centering
        \includegraphics[width=\linewidth]{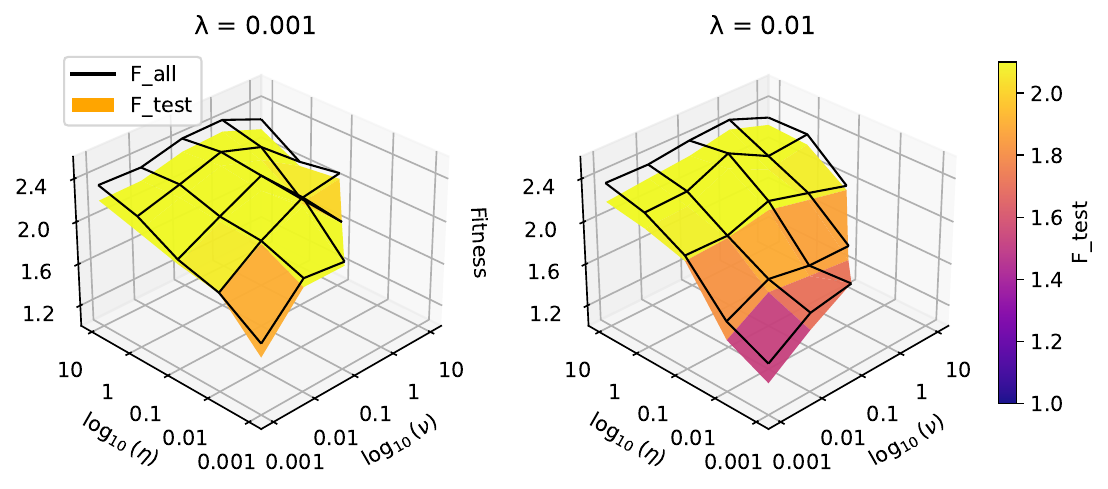}
        \caption{\textcolor{black}{Diversity-DeepSAD using HT features.}}
        \label{fig:deepsad_ht_sensitivity_reduced}
    \end{subfigure}

    \caption{\textcolor{black}{Hyperparameter sensitivity analysis of augmented Diversity-DeepSAD over $(\nu, \eta, \lambda)$, for the two SP methods.}}
    \label{fig:deepsad_sensitivity_reduced}
\end{figure}

\textcolor{black}{Regarding Diversity-DeepSAD, the surfaces $F_{all}(\eta,\nu|\lambda)$ and $F_{test}(\eta,\nu|\lambda)$ for $\lambda$ of 0.001 and 0.01 in \autoref{fig:deepsad_sensitivity_reduced}, with full plots in \ref{app: sensitivity}, revealed that performance for both features is governed primarily by the diversity weight $\lambda$, with some sensitivity to the auxiliary label weight $\eta$ and weaker dependence on the regularisation weight $\nu$. For small values of $\lambda$, a plateau of high fitness values was observed, with $F_{all}$ and $F_{test}$ remaining close to their optima. Only very large $\nu$ values systematically degraded performance, showing that the regularisation term can have a significant effect. The large dependence on $\lambda$ confirmed that the diversity term is essential for producing smooth, monotonic HIs, while the regularisation refined performance. Overall, the configuration used ($\eta  = 10, \ \nu = 10, \ \lambda = 0.001 $) lies well within the stable high-performance region and is therefore fully justified.}

\begin{figure}[H]
    \centering

    \begin{subfigure}{0.70\linewidth}
        \centering
        \includegraphics[width=\linewidth]{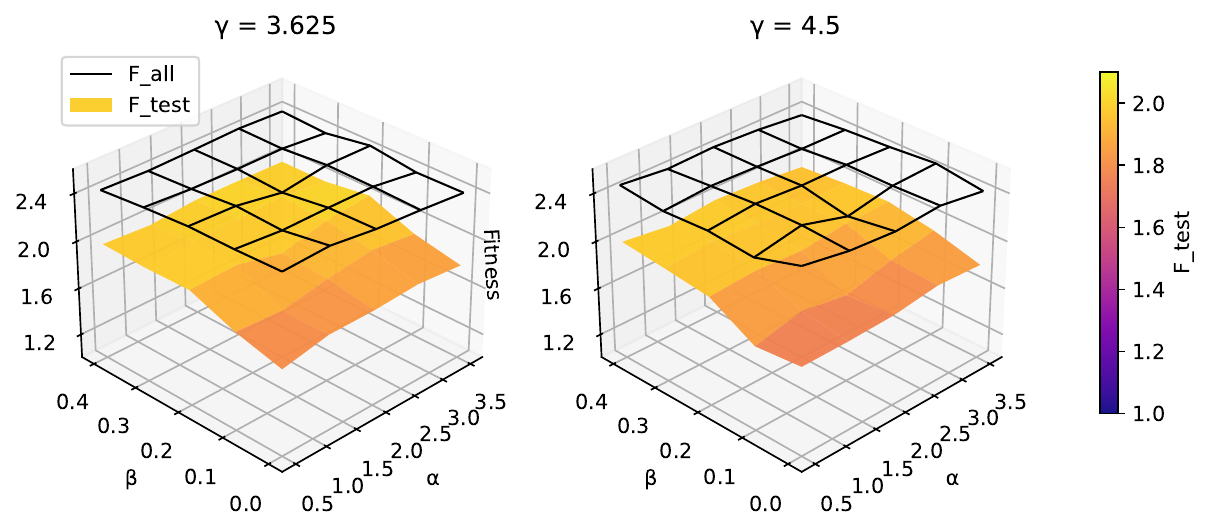}
        \caption{\textcolor{black}{DTC-VAE using FFT features.}}
        \label{fig:vae_fft_sensitivity_reduced}
    \end{subfigure}

    \vspace{8pt}

    \begin{subfigure}{0.70\linewidth}
        \centering
        \includegraphics[width=\linewidth]{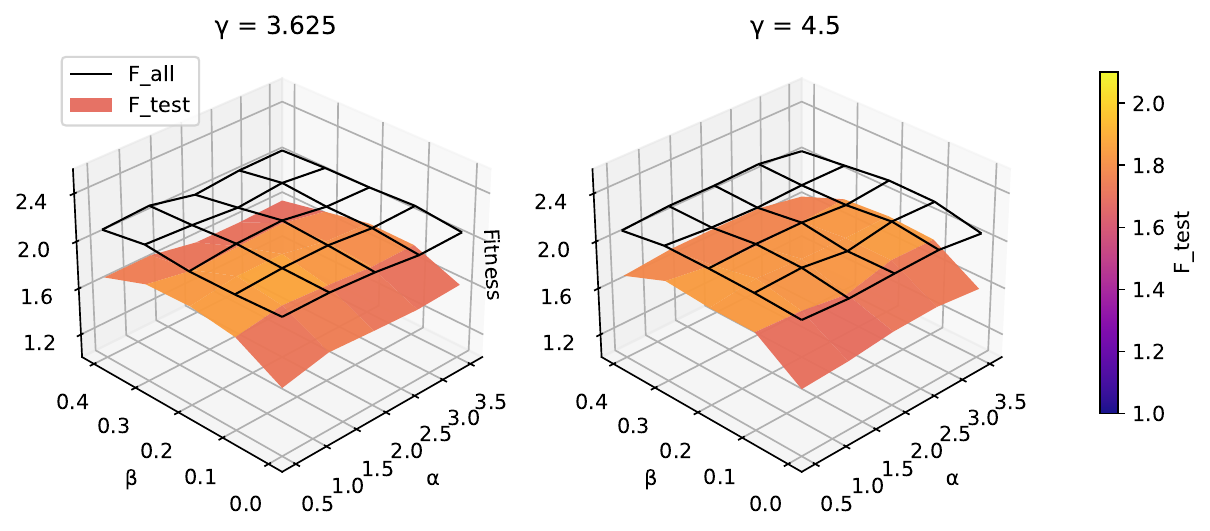}
        \caption{\textcolor{black}{DTC-VAE using HT features.}}
        \label{fig:vae_ht_sensitivity_reduced}
    \end{subfigure}

    \caption{\textcolor{black}{Hyperparameter sensitivity analysis of DTC-VAE over $(\alpha, \beta, \gamma)$, for the two SP methods.}}
    \label{fig:vae_sensitivity_reduced}
\end{figure}

\textcolor{black}{Regarding DTC-VAE, the response surfaces $F_{all}(\alpha,\beta|\gamma)$ and $F_{test}(\alpha,\beta|\gamma)$ exhibited a stronger stability across all hyperparameters, with two examples shown in \autoref{fig:vae_sensitivity_reduced} and all figures being included in \ref{app: sensitivity}. Increases beyond the optimised regions did lead to gradual changes in fitness. Very small $\gamma$ values saw the greatest decrease in fitness, highlighting the importance of the monotonicity constraint in constructing reliable HIs. Furthermore, for small $\gamma$, increases in $\beta$ beyond the optimised region led to reduced fitness, consistent with an overly strong reconstruction loss term dominating the objective, reducing the effective influence of the KL term and thus degrading the learned latent structure. However, this was not observed when the monotonicity constraint was dominant, showing improved stability. In all cases, the ranges optimised in the main experiments fell inside a stable high-fitness region, rather than at a sharp optimum.}

\textcolor{black}{Overall, the sensitivity analysis shows that both models maintain high $F_{all}$ and $F_{test}$ scores across a non-trivial neighbourhood of their nominal hyperparameters. This demonstrates that the comparative advantages of FFT over HT, and of DTC-VAE over Diversity-DeepSAD, arise from genuinely robust model behaviour rather than from narrowly tuned hyperparameter settings.}

\vspace{2mm}

\section{Discussion}\label{sec:disc}

\noindent \textcolor{black}{This section contextualises the results by analysing the behaviour of the constructed HIs, assessing their robustness and computational feasibility, and comparing the proposed models with state-of-the-art methods. The discussion links model behaviour to underlying physical mechanisms and highlights practical implications for GW SHM.}

\vspace{-7mm}

\textcolor{black}{\subsection{Interpretation of constructed HIs}}

\noindent \textcolor{black}{Developing robust and reliable HIs for composite structures remains a challenging task, as these indicators must capture meaningful trends while remaining interpretable and suitable for prognostic applications.} \textcolor{black}{This analysis evaluates the augmented Diversity-DeepSAD and DTC-VAE, highlighting key differences in their performances and behaviours, which reveal their} respective strengths and limitations in extracting meaningful HIs for composite structure prognostics.

Diversity-DeepSAD graphs show a tendency for HIs to increase in sharp steps, which could indicate different damage states during the fatigue life of composite structures. However, these increases were sometimes erroneous or abrupt; this is most visible in \autoref{fig:DeepSAD_HLB} in test Sample 3 at \SI{50}{\kilo\hertz} and test Sample 4 at \SI{125}{\kilo\hertz}, showing the model may be sensitive to sudden changes in test sample structural health or sensor damage. On the other hand, DTC-VAE produced smoother HIs, indicating improved performance in this regard. This is particularly the case with FFT features, with some exceptions in the case of HT features.

\textcolor{black}{The incremental changes in HIs produced by Diversity-DeepSAD could be seen as a more explainable result, potentially illustrating the distinct damage states of composite structures. This could provide valuable insights for prognostic models, especially state-based ones, in predicting RUL. However, the increases in HIs should not be too abrupt, else using HIs for RUL prediction would be challenging, as discussed previously. Further studies and testing are needed to establish a stronger connection between these phases and actual physical damage states. On the other hand, HIs produced by DTC-VAE resulted in slightly higher fitness scores and less sharp jumps, that could improve RUL prediction.}

\begin{figure}[h]
    \centering
    \includegraphics[width=0.8\textwidth]{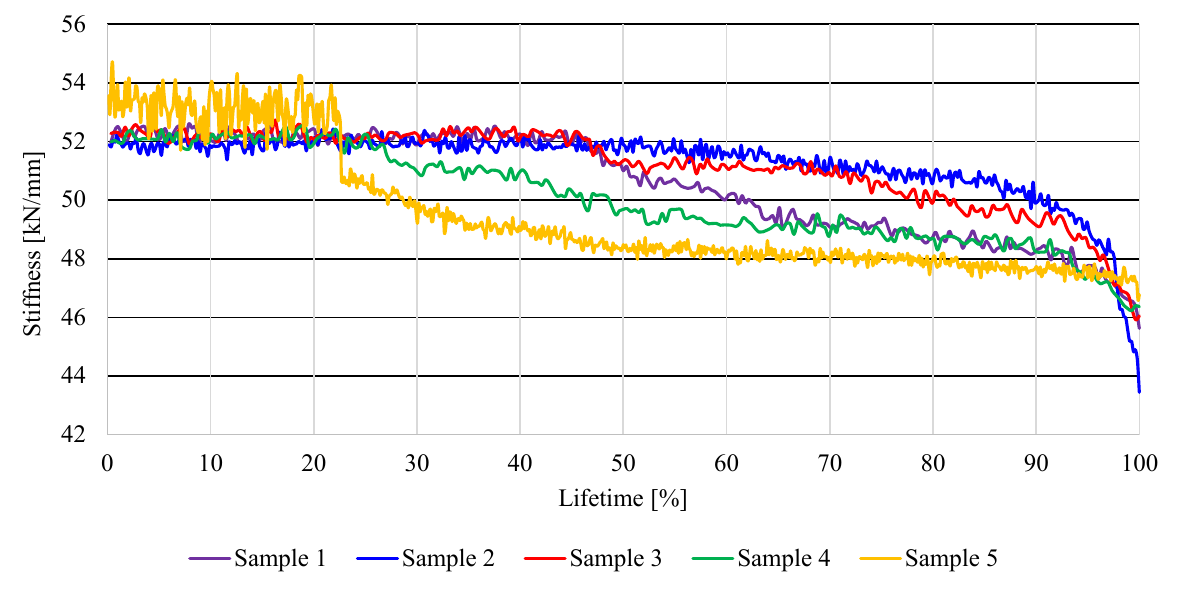}
    \caption{\textcolor{black}{Post-buckling stiffness against number of compression cycles for each specimen.}}
    \label{fig:stresscyclegraph}
\end{figure}

\textcolor{black}{In order for further physical interpretation, \autoref{fig:stresscyclegraph} shows the stiffness of each specimen, which is estimated by the slope of the linear region of the load–displacement curves during post-buckling (i.e.\ ranging from 40 kN to 60 kN) \cite{Yue2022}. The stiffness reduction is due to complex damage mechanisms and therefore also represents the physical health of the structure, similarly to the generated HIs. It should be noted that stiffness was not included in the data fed to the models, as it cannot be easily measured in operation, particularly in an aerospace application.}

\textcolor{black}{The HIs produced by augmented Diversity-DeepSAD and DTC-VAE can be meaningfully linked to plausible damage mechanisms and measured stiffness trends, while remaining cautious about causality (see \autoref{tab:HIbehaviour}). Stepwise HIs produced by Diversity-DeepSAD highlight discrete changes in GW scattering\textemdash consistent with localised events such as delamination initiation, disbond propagation, or sudden coalescence of cracks\textemdash whereas the DTC-VAE yields smoother HIs that reflect the gradual accumulation of distributed microdamage (e.g.\ matrix cracking). Considering the impact/disbond locations, that of Sample 2 lies off the stiffener and shows the earliest HI transition but the smoothest stiffness decline, suggesting local damage detected early by GW but not yet affecting global stiffness; Sample 4's is located on the stiffener and exhibits long stability followed by late HI jumps and a sharper stiffness loss, consistent with critical stiffener-related propagation; Sample 5’s pre-existing disbond explains its early, abrupt stiffness reduction ($\sim$20\% lifetime) while GW responses are path-dependent and thus more model-sensitive. Where both models and feature folds consistently indicate transitions (for instance Sample 2’s early change), confidence is increased that the HI reflects a genuine physical change. Differences in timing between HI transitions and stiffness drops therefore provide diagnostic value: early HI changes with late stiffness loss point to local damage not yet structural, while early stiffness drops with variable GW response point to gross load-path alterations (e.g.\ a disbond).}

\begin{table}[!tbh]
\centering
\caption{\textcolor{black}{Summary of data-driven HI behaviour (constructed by augmented Diversity-DeepSAD vs.\ DTC-VAE, both upon FFT features), stiffness evolution, impact/disbond location, and plausible physical interpretation.}}
\vspace{-5pt}
\resizebox{\textwidth}{!}{%
\begin{tabular}{l|llll|l}
\hline
\textbf{Specimen} & \textbf{\begin{tabular}[c]{@{}l@{}}Impact/disbond \\ \& relative position\end{tabular}} & \textbf{\begin{tabular}[c]{@{}l@{}}DeepSAD's HI \\ (damage states before \\ final jump to EoL \& timing)\end{tabular}} & \textbf{\begin{tabular}[c]{@{}l@{}}DTC-VAE's HI \\ (shape \& timing)\end{tabular}} & \textbf{Stiffness behaviour} & \textbf{Plausible physical interpretation (linking HI → stiffness)} \\ \hline
\textbf{Sample 1} & \begin{tabular}[c]{@{}l@{}}Impact near \\ stiffener edge\end{tabular} & \begin{tabular}[c]{@{}l@{}}2 damage states; \\ first state longer than second\end{tabular} & \begin{tabular}[c]{@{}l@{}}Very smooth HI; \\ gradual monotonic rise\end{tabular} & \begin{tabular}[c]{@{}l@{}}Smoothed sudden \\ reduction, later in \\ life\end{tabular} & \begin{tabular}[c]{@{}l@{}}Impact near stiffener edge causes early local cracking/small \\ delaminations detected by GW as two discrete stages (DeepSAD). \\ Global stiffness declines later when local damage coalesces. \\ DTC-VAE’s smooth HI captures overall monotonic accumulation.\end{tabular} \\ \hline
\textbf{Sample 2} & \begin{tabular}[c]{@{}l@{}}Impact at \\ off-stiffener region\end{tabular} & \begin{tabular}[c]{@{}l@{}}3 damage states; \\ earliest first transition \\ among samples; \\ long final state\end{tabular} & \begin{tabular}[c]{@{}l@{}}Fast HI rise early \\ (first $\sim$30\% lifetime), \\ then slow growth\end{tabular} & \begin{tabular}[c]{@{}l@{}}Smoothest stiffness \\ decline; no early \\ jump (stiffness \\ accelerates near \\ EoL)\end{tabular} & \begin{tabular}[c]{@{}l@{}}Off-stiffener impact produces local scattering changes that GW \\ detects very early (DeepSAD). Because damage is local and not \\ initially in primary load paths, global stiffness remains unaffected \\ until late accumulation—consistent with DTC-VAE plateau after \\ early rise.\end{tabular} \\ \hline
\textbf{Sample 3} & \begin{tabular}[c]{@{}l@{}}Impact near \\ opposite stiffener edge\end{tabular} & \begin{tabular}[c]{@{}l@{}}4 damage states; \\ 3rd state longest\end{tabular} & \begin{tabular}[c]{@{}l@{}}Gradual multi-stage \\ trend, intermediate \\ between extremes\end{tabular} & \begin{tabular}[c]{@{}l@{}}Smooth reduction; \\ late acceleration \\ toward EoL\end{tabular} & \begin{tabular}[c]{@{}l@{}}Multiple HI states suggest sequential damage mechanisms (matrix \\ cracking → interface debonding → delamination) at/around the \\ stiffener edge. Stiffness reduces progressively and accelerates as \\ damage becomes global.\end{tabular} \\ \hline
\textbf{Sample 4} & \begin{tabular}[c]{@{}l@{}}Impact on \\ the stiffener, center\end{tabular} & \begin{tabular}[c]{@{}l@{}}2 damage states;\\ fewer pre-final stages\end{tabular} & \begin{tabular}[c]{@{}l@{}}Slow progression until \\ $\sim$70\% lifetime, then jump(s) \\ toward EoL\end{tabular} & \begin{tabular}[c]{@{}l@{}}Early-to-mid life \\ smoother drop \\ then sharper loss\end{tabular} & \begin{tabular}[c]{@{}l@{}}Impact on stiffener produces damage that remains subcritical for \\ long time (little HI change) then propagates rapidly along the stiffener \\ causing HI jumps and stiffness loss (loss of load-transfer at stiffener). \\ Timing matches DTC-VAE late jumps.\end{tabular} \\ \hline
\textbf{Sample 5} & \begin{tabular}[c]{@{}l@{}}Manufacturing disbond \\ between stiffener and \\ skin (right side)\end{tabular} & \begin{tabular}[c]{@{}l@{}}2 damage states (sometimes \\ 3 depending on fold); \\ early transition visible\end{tabular} & \begin{tabular}[c]{@{}l@{}}Smooth monotonic HI \\ but early rise relative \\ to some units\end{tabular} & \begin{tabular}[c]{@{}l@{}}Earliest stiffness \\ decrease \\ ($\sim$20\% lifetime); \\ sudden early drop\end{tabular} & \begin{tabular}[c]{@{}l@{}}Pre-existing disbond alters global load path → early measurable \\ stiffness loss. GW HIs detect local scattering but effectiveness depends\\  on which paths intersect the disbond; model variability reflects path \\ sensitivity. Overall, early stiffness drop is consistent with a structural \\ discontinuity (disbond) rather than purely fatigue accumulation.\end{tabular} \\ \hline
\end{tabular}
}\label{tab:HIbehaviour}
\end{table}

Furthermore, results from both DTC-VAE and in particular Diversity-DeepSAD showed a tendency for HIs to spike at EoL, such as test Sample 3 at \SI{125}{\kilo\hertz}. This is due to the model learning the failure state as a separate category to the continuous range of degradation until that point, an effect which can be emphasised with alternative hyperparameters. This suggests that the deviation in the structural health condition after failure is significantly greater and distinct from the condition before failure, demonstrating the damage accumulation process in composite structures. Although high prognosability scores may indicate that extracted HIs are more suitable for RUL prediction, it is crucial to ensure that changes in HI values are not too abrupt. For example, a sharp increase near the end of life may leave insufficient time to take maintenance action. In this regard, DTC-VAE could potentially provide HIs that are more useful than Diversity-DeepSAD.

The improved performance of DTC-VAE can be linked to the connection between its loss function and the fitness of the extracted HIs. The incorporation of the monotonicity constraint into its loss function makes part of the fitness metric differentiable, and thus, optimisable. This provides a direct relationship between training and output fitness, by penalizing the model for producing HIs with lower fitness scores, ensuring that these follow a smooth, monotonic trend. This trend is easier to predict, making the extracted HIs more suitable for prognostics.

\vspace{-7mm}

\textcolor{black}{\subsection{Uncertainty and dataset size}}

\noindent \textcolor{black}{The suitability of incorporating uncertainty quantification into the evaluation of the constructed HIs was examined, given its common application for prognostic modelling. Although it is widely used in RUL prediction \cite{Wang2024UC}, its use in HI construction remains limited. Studies on HIs typically rely on the use of prognostic criteria and do not report confidence intervals \cite{Moradi2024, Yang2023}. This is partly because HIs represent abstract latent variables rather than directly measurable physical quantities. As a result, interval-based uncertainty estimates for fitness scores may be misleading, especially in relation to smaller datasets where statistical variance cannot be reliably defined. Nevertheless, as reported in this study, evaluating the standard deviation of $Mo$, $Pr$, $Tr$, and the fitness score\textemdash across (i) multiple random initialisations of the deep learning models, which reflects model uncertainty, and (ii) different composite specimens, which reflects variability due to GW sensing, manufacturing imperfections, and minor differences in sensor placement and test setup\textemdash is necessary. These statistics provide a quantitative assessment of the robustness of the learned HIs and their sensitivity to both model and experimental variability.}

\textcolor{black}{Although only five composite specimens were employed in this study, the dataset is highly information-rich: each unit provides dozens of GW measurements, each comprising 56 actuator–sensor paths across six frequencies. Since the proposed models operate in a history-independent manner, every GW measurement timestep forms an independent sample, yielding approximately 230 machine learning measurements in total. Furthermore, the leave-one-unit-out evaluation, where an entire specimen is unseen during training, provides a stringent generalisation test and helps mitigate overfitting despite the limited number of structural units.}

\vspace{0mm}

\textcolor{black}{\subsection{Computational feasibility}}

\noindent \textcolor{black}{Both Diversity-DeepSAD and DTC-VAE remain computationally lightweight. All experiments were conducted on a laptop equipped with a 12th-Gen Intel Core i7-12700H CPU (14 cores, 20 threads), 16 GB RAM and an NVIDIA RTX A1000 Laptop GPU with 4 GB VRAM. Training and inference were performed using GPU acceleration (CUDA 12.8). Training a single model instance, in other words one frequency and feature type for one validation fold, required on average \SI{35.68}{\second} for Diversity-DeepSAD and \SI{2.27}{\second} for DTC-VAE, with a maximum of \SI{62.34}{\second} and \SI{3.01}{\second} respectively. The corresponding parameter counts averaged 519,664 trainable parameters for Diversity-DeepSAD and 71,309 for DTC-VAE (1.982 MB and 0.272 MB respectively in memory), depending mainly on the input dimensionality and hidden layer width.}

\textcolor{black}{Inference for a new specimen, i.e.\ generating a complete HI trajectory for a single specimen at one frequency, took on average \SI{17.47}{\milli\second} per forward pass for Diversity DeepSAD and 1.03~ms for DTC-VAE. Since training and Bayesian optimisation are performed offline, only this inference step is relevant for online SHM. The sub-second inference times leave a large margin with respect to typical GW measurement intervals, indicating that both methods are suitable for real-time deployment, for example, for in-service aircraft structures in the industrial context. Moreover, the training time scales approximately linearly with the number of specimens and epochs, and model size scales linearly with the hidden layer width, suggesting that the framework can be extended to larger datasets and additional frequencies without prohibitive computational overhead.}

\vspace{-3mm}

\textcolor{black}{\subsection{Comparison with state-of-the-art}}

\noindent Based on the average prognostic criteria and fitness scores across all units, Diversity-DeepSAD and DTC-VAE demonstrate competitive performance compared to other proposed models, as shown in \autoref{tab:FitnessXotherLiterature}. In this table, two SHM techniques—GW and acoustic emission—are also compared. It should be noted that the constructed HIs using acoustic emission are inherently time-dependent due to the nature of the sensing technology, whereas the GW technique has the potential to provide data for designing history-independent HIs, as demonstrated by the proposed models in this study and HT-SSCNN \cite{Moradi2024}.

While SSLSTM using acoustic emission data \cite{Moradi2023Intelligent} achieves the highest fitness score across all units ($F_{all}$) with a performance of 93\%, it relies on test units for validation during the LOOCV process to stop training and fine-tune hyperparameters. This reliance potentially limits its applicability to real-world scenarios where test units are new and unknown. However, this limitation has been addressed in subsequent work \cite{moradi2025novel}, where CEEMDAN-SSLSTM followed by a BiLSTM ensemble achieved the best performance for acoustic emission data, with 91.3\% for $F_{all}$ and 86.3\% for test units ($F_{test}$).

On the other hand, using GW data, the proposed history-independent HIs in this study result in the lowest deviation between fitness scores obtained for all units ($F_{all}$) and test units ($F_{test}$), demonstrating reduced overfitting. Both the semi-supervised Diversity-DeepSAD model and the unsupervised DTC-VAE model are significantly more stable, with much lower standard deviations compared to existing state-of-the-art models. The DTC-VAE model, followed by WAE fusion, achieved the highest performance of 92\% for test units ($F_{test}$) and 92.3\% for all units ($F_{all}$). Notably, the deviation between fitness scores for all units and test units is minimal (0.3\%).

\begin{table}[!tbh]
\centering
\caption{Performance of various methods in the ReMAP dataset collected using guided wave (GW) and acoustic emission techniques.}
\vspace{-5pt}
\resizebox{\textwidth}{!}{%
\begin{tabular}{c|ccc|cc}
\toprule
\textbf{\begin{tabular}[c]{@{}c@{}}Prognostic\\Criteria\end{tabular}} & \textbf{\begin{tabular}[c]{@{}c@{}}Diversity-DeepSAD\\on FFT (WAE)\end{tabular}} & \textbf{\begin{tabular}[c]{@{}c@{}}DTC-VAE\\on FFT (WAE)\end{tabular}} & \textbf{\begin{tabular}[c]{@{}c@{}}HT-SSCNN \cite{Moradi2024}\\(WAE)\end{tabular}} & \textbf{\begin{tabular}[c]{@{}c@{}}SSLSTM \cite{Moradi2023Intelligent}\\(simple ensemble)\end{tabular}} & \textbf{\begin{tabular}[c]{@{}c@{}}CEEMDAN-SSLSTM \cite{moradi2025novel}\\(BiLSTM ensemble)\end{tabular}}\\ \midrule
$\boldsymbol{Mo}$  & 0.94 (\textpm 0.01)   & 0.99 (\textpm 0.01) & - & 0.99 (\textpm 0.01) & -   \\ 

$\boldsymbol{Pr}$   & 0.91  (\textpm 0.03)  & 0.98 (\textpm 0.01)  & - & 0.86 (\textpm 0.14) & -  \\ 

$\boldsymbol{Tr}$  & 0.60 (\textpm 0.08) & 0.80 (\textpm 0.05)      & - & 0.93 (\textpm 0.03) & - \\ 

$\boldsymbol{\textbf{\textit{$F_{all}$}}}$  & 2.45 (\textpm 0.06) & 2.77 (\textpm 0.05)  & 2.78 (\textpm 0.15) & 2.79 (\textpm 0.14) & 2.74 (\textpm 0.19) \\ 

\midrule
$\boldsymbol{Mo_{test}}$  & 0.93 (\textpm 0.04)   & 0.98 (\textpm 0.02) & - & -   & - \\ 

$\boldsymbol{Pr_{test}}$   & 0.86  (\textpm 0.10)  & 0.97 (\textpm 0.02)  & - & - & - \\ 

$\boldsymbol{Tr}$  & 0.60 (\textpm 0.08) & 0.80 (\textpm 0.05)      & - & - & - \\ 

$\boldsymbol{\textbf{\textit{$F_{test}$}}}$  & 2.39 (\textpm 0.12) & 2.76 (\textpm 0.06)  & 2.67 (\textpm 0.20) & - & 2.59 (\textpm 0.24) \\ 

\midrule
\multicolumn{1}{l}{\textit{Number of units}}  & \multicolumn{3}{c}{\textit{5 composite specimens}}  & \multicolumn{2}{c}{\textit{12 composite specimens}}   \\

\multicolumn{1}{l}{\textit{SHM technique}}   & \multicolumn{3}{c}{\textit{GW}} & \multicolumn{2}{c}{\textit{Acoustic Emission}}   \\ 
\bottomrule                                                    
\end{tabular}%
}\label{tab:FitnessXotherLiterature}
\end{table}

\textcolor{black}{Although this study focuses on the model families previously explored for the ReMAP GW dataset, the results highlight opportunities for evaluating more advanced architectures. Models such as transformers, graph neural networks, and physics-informed neural networks may further exploit spatial–temporal dependencies in GW signals. A systematic assessment of these architectures is left for future work.}
\vspace{-2mm}

\textcolor{black}{\subsection{Limitations and future work}}

\noindent \textcolor{black}{Although the proposed framework demonstrates strong performance and robustness, some limitations identified in the discussion are summarised explicitly:
\begin{enumerate}[label=(\roman*)]
    \item The dataset contains only five composite specimens. The data richness and leave-one-unit-out evaluation provide sufficient validation coverage, however, future work should aim to include more specimens to improve statistical generalisation.
    \item Diversity-DeepSAD and DTC-VAE occasionally produce abrupt transitions at EoL, somewhat influencing the prognosability criterion and their use for prognostics. Further work should aim to incorporate further constraints into the learning process and loss functions.
    \item The framework relies on multi-frequency guided wave measurements with dense actuator–sensor paths, which may not be available in all SHM deployments.
    \item Only two model families were explored. More advanced architectures, such as transformers, graph neural networks, and physics-informed neural networks, may further improve degradation representation.
\end{enumerate}}

\vspace{-5mm}

\section{Conclusion}\label{sec:conc}

\vspace{-3mm}

\noindent This study proposed a novel plenary data-driven framework for constructing health indicators (HIs) for aeronautical composite structures by developing two advanced machine learning models: Diversity-DeepSAD and DTC-VAE. First, statistical features from the best-performing signal processing techniques, fast Fourier transform (FFT) and Hilbert transform (HT), were extracted as inputs for the AI models. This step enabled dimensionality reduction of GW signals, each 2000 data points in length, recorded by PZTs across 56 actuator-sensor inspection paths at six excitation frequencies (2000×56×6 data points per measurement), to just a few tens of features per excitation frequency. This allows for simpler deep learning architectures while preserving critical information. Feature selection was based on fitness scores, calculated as the sum of the prognostic criteria of monotonicity, prognosability, and trendability. FFT features demonstrated superior performance, achieving a performance of 71.0\%, i.e.\ an average fitness score of 2.13 (out of 3.00), compared to 68.2\% for HT features.

Diversity-DeepSAD and DTC-VAE exhibited distinct \textcolor{black}{behaviours} in HI construction. Diversity-DeepSAD produced incremental changes in HIs, potentially reflecting distinct damage states, but occasionally showed abrupt increases, particularly near the end of life (EoL). In contrast, DTC-VAE generated smoother HIs with higher fitness scores, benefiting from its monotonicity-constrained loss function, which directly optimised the fitness metric. This smoother trend makes DTC-VAE more suitable for RUL prediction, as it avoids abrupt changes that could limit actionable maintenance time. Therefore, the augmented Diversity-DeepSAD is recommended as a closer representative of true health state, while DTC-VAE may be more useful for prognostic applications.

The proposed framework outperformed state-of-the-art models in constructing history-independent HIs. DTC-VAE, followed by weighted averaging ensemble fusion, achieved the highest test fitness score of 2.76 (92\% performance) with minimal deviation (0.3\%) from the fitness score across all units, indicating reduced overfitting and improved \textcolor{black}{generalisation}. Moreover, the models demonstrate greater stability in performance compared to existing approaches. Unlike some existing models, which rely on test units for validation, the proposed framework ensures applicability to real-world scenarios where test units are unknown.
 
Overall, this study \textcolor{black}{addresses the critical challenge of constructing robust HIs for aeronautical composite structures, where reliable monitoring and prognostics are essential for operational safety and effective maintenance planning.} \textcolor{black}{By leveraging the strengths of augmented Diversity-DeepSAD and DTC-VAE, the proposed frameworks provide stable HIs} while supporting more reliable RUL predictions. \textcolor{black}{The frameworks are designed to be computationally efficient, scalable and feasible for monitoring, making them suitable for SHM applications such as in-service aircraft structures.} Future work will focus on extending the DTC-VAE loss function to incorporate additional prognostic criteria, \textcolor{black}{exploring larger datasets with more units to validate generalisability and assessing the implementation in operational environments to further enhance the frameworks practical} applicability. \textcolor{black}{Additional work will also explore the applicability of graph neural networks and physics-informed neural networks to GW SHM.}

\bibliographystyle{elsarticle-num}
\bibliography{Bibliography}

\appendix
\section{List of features by domain}\label{app:A}

\noindent \textcolor{black}{Each time or time-frequency domain feature $X$ is given in terms of the time or time-frequency domain signal samples $x$, while frequency domain features $S$ are given in terms of the frequency domain signal samples $s(f)$,}

\begin{table}[H]
        \centering
        \caption{Time domain features \cite{Moradi2023Intelligent}.}
        \label{tab:timedomainfeatures}
        \makebox[0.8\textwidth]{
        {\def\arraystretch{2}
        \begin{tabular}{l c c |l c c} \hline
           \textbf{No} & Feature & Equation & \textbf{No} & Feature & Equation \\ \hline
           \textbf{1} & Mean value & $X_m = \frac{\Sigma^{N}_{i=1} x(i)}{N}$ & \textbf{11} & Shape factor &  $X_{shape} = \frac{X_{rms}}{\frac{1}{N}\Sigma^{N}_{i=1} \lvert x(i) \rvert}$\\ 
           \textbf{2} & Standard deviation & $X_{sd} = \sqrt{\frac{\Sigma^{N}_{i=1} (x(i)-X_m)^2}{N-1}}$ & \textbf{12} & Impulse factor & $X_{impulse} = \frac{X_{peak}}{\frac{1}{N}\Sigma^{N}_{i=1} \lvert x(i) \rvert}$\\ 
           \textbf{3} & Root amplitude & $X_{root} = \left(\frac{\Sigma^{N}_{i=1} \sqrt{\lvert x(i) \rvert}}{N} \right)^2$ & \textbf{13} & \makecell{Max-min \\ difference} &
           \makecell{$X_{p2p}=max\left(x(i)\right)$ \\ \hspace{1cm}$-min\left(x(i)\right)$}\\
            
           \textbf{4} & Root mean square & $X_{rms} = \sqrt{\frac{\Sigma^{N}_{i=1} (x(i))^2}{N}}$ & \multirow{2}{*} {\makecell{\textbf{14}\\\textbf{-17}}} & \multirow{2}{*} {\makecell{$k$th central \\ moment \\ ($k=3, 4, 5, 6$)}} & \multirow{2}{*}{$X_{k\_cm} = \frac{\Sigma^{N}_{i=1} \left(x(i)- X_m\right)^k}{N}$}  \\
           
           \textbf{5} & \makecell{Residual sum of \\ squares} & $X_{rss} = \sqrt{\Sigma_{i=1}^N \lvert x(i) \rvert^2}$ \\ 
           \textbf{6} & Peak maximum & $X_{peak} = max(|x(i)|)$ & \textbf{18} & FM4 & $X_{FM4} = \frac{X_{4\_cm}}{X_{sd}^4}$ \\   
           \textbf{7} & Skewness & $X_{skewness} = \frac{\Sigma^{N}_{i=1} \left(x(i)- X_m\right)^3}{(N-1) X_{sd}^3}$ &  \textbf{19} & Median & $X_{med} = \frac{\Sigma^{N}_{i=1} t(i)}{N}$\\    
           \textbf{8} & Kurtosis & $X_{kurtosis} = \frac{\Sigma^{N}_{i=1} \left(x(i)- X_m\right)^4}{(N-1) X_{sd}^4}$ & &&\\    
           \textbf{9} & Crest factor & $X_{crest} = \frac{X_{peak}}{X_{rms}}$& & &\\   
           \textbf{10} & Clearance factor & $X_{clearance} = \frac{X_{peak}}{X_{root}}$ &  && \\[0.3cm]
           \hline

        \end{tabular}}}
\end{table}

\begin{table}[H]
        \centering
        \caption{Frequency domain features \textcolor{black}{$S$} \cite{Moradi2023Intelligent}.}
        \label{tab:freqdomainfeatures}
        \makebox[0.8\textwidth]{
        {\def\arraystretch{2}
        \begin{tabular}{l c c |l c c} \hline
           \textbf{No} & Feature & Equation & \textbf{No} & Feature & Equation \\ \hline
           & & & & & \\[-0.5cm]
           \textbf{1} & Mean frequency & $S_1 = X_{mf} = \frac{\Sigma^{N_f}_{f=1} s(f)}{N_f}$ & \textbf{8} & - & $S_8 = \sqrt{\frac{\Sigma^{N_f}_{f=1} f_k^4s(f)}{\Sigma^{N_f}_{f=1} f_k^2s(f)}}$\\ 
           \textbf{2} & \textit{(same as variance)} & $S_2 = \frac{\Sigma^{N_f}_{f=1} \left(s(f)-S_1\right)^2}{N_f-1}$ & \textbf{9} & - & $S_9 = \frac{\Sigma_{f=1}^{N_{f}} f_k^2 s(f)}{\sqrt{\Sigma_{f=1}^{N_{f}} s(f) \Sigma_{f=1}^{N_{f}} f_k^4}}$ \\ 
           \textbf{3} & \textit{(same as skewness)} & $S_3 = \frac{\Sigma^{N_f}_{f=1} \left(s(f)-S_1\right)^3}{N_f \left(\sqrt{S_2}\right)^3}$ & \textbf{10} & - & $S_{10} = \frac{S_6}{S_5}$\\ 
           & & & & & \\[-0.7cm]
           \textbf{4} & \textit{(same as kurtosis)} & $S_4 = \frac{\Sigma^{N_f}_{f=1} \left(s(f)-S_1\right)^4}{N_f S_2^2}$ & \textbf{11} & - & $S_{11} = \frac{\Sigma_{f=1}^{N_{f}} (f_k - S_5)^3 s(f)}{N_f S_6^3}$\\ 
           \textbf{5} & -  & $S_5 = X_{fc} = \frac{\Sigma_{f=1}^{N_{f}} f_k s(f)}{\Sigma_{f=1}^{N_{f}} s(f)}$ & \textbf{12} & - & $S_{12} = \frac{\Sigma_{f=1}^{N_{f}} (f_k - S_5)^4 s(f)}{N_f S_6^4}$\\ 
           & & & & & \\[-0.7cm]
           \textbf{6} & - & $S_6 = \sqrt{\frac{\Sigma_{f=1}^{N_{f}}(f_k-S_5)^2 s(f)}{N_f}}$& \textbf{13} & - & $S_{13} = \frac{\Sigma_{f=1}^{N_{f}} \sqrt{(f_k - S_5)} s(f)}{N_f \sqrt{S_6}}$\\ 
           \textbf{7} & - & $S_7 = X_{rmsf} = \sqrt{\frac{\Sigma_{f=1}^{N_{f}} f_k^2 s(f)}{\Sigma_{f=1}^{N_{f}} s(f)}}$& \textbf{14} &  - & $S_{14} = \sqrt{\frac{\Sigma_{f=1}^{N_{f}} (f_k-S_5)^2 s(f)}{\Sigma_{f=1}^{N_{f}} s(f)}}$\\[0.4cm]
           \hline

        \end{tabular}}}
\end{table}

\begin{table}[H]
        \centering
        \caption{Time-frequency domain features \textcolor{black}{$X$} \cite{Buckley2023}.}
        \label{tab:STFTdomain}
        {\def\arraystretch{2}
        \begin{tabular}{l c c } \hline
           \textbf{No} & Feature & Equation  \\ \hline
           \textbf{1} & Mean value & $X_m = \frac{\Sigma^{N}_{i=1} x(i)}{N}$ \\ 
           \textbf{2} & Standard deviation & $X_{sd} = \sqrt{\frac{\Sigma^{N}_{i=1} (x(i)-X_m)^2}{N-1}}$ \\
           \textbf{3} & Skewness & $X_{skewness} = \frac{\Sigma^{N}_{i=1} \left(x(i)- X_m\right)^3}{(N-1) X_{sd}^3}$\\ 
           \textbf{4} & Kurtosis & $X_{kurtosis} = \frac{\Sigma^{N}_{i=1} \left(x(i)- X_m\right)^4}{(N-1) X_{sd}^4}$ \\[0.4cm]
           \hline

        \end{tabular}}
\end{table}
\section{\textcolor{black}{Sensitivity Analysis}} \label{app: sensitivity}
\begin{figure}[H]
    \centering

    \begin{subfigure}{\linewidth}
        \centering
        \includegraphics[width=\linewidth]{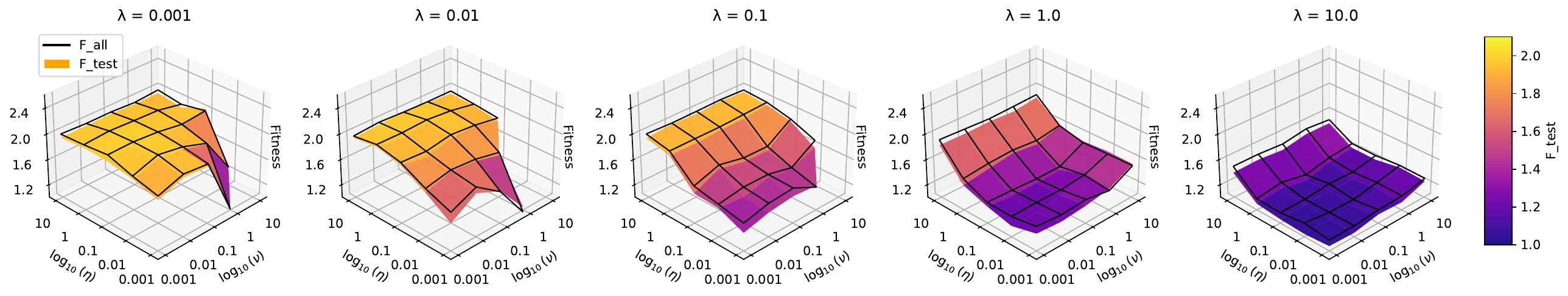}
        \caption{\textcolor{black}{Diversity-DeepSAD using FFT features.}}
        \label{fig:deepsad_fft_sensitivity}
    \end{subfigure}

    \vspace{8pt}

    \begin{subfigure}{\linewidth}
        \centering
        \includegraphics[width=\linewidth]{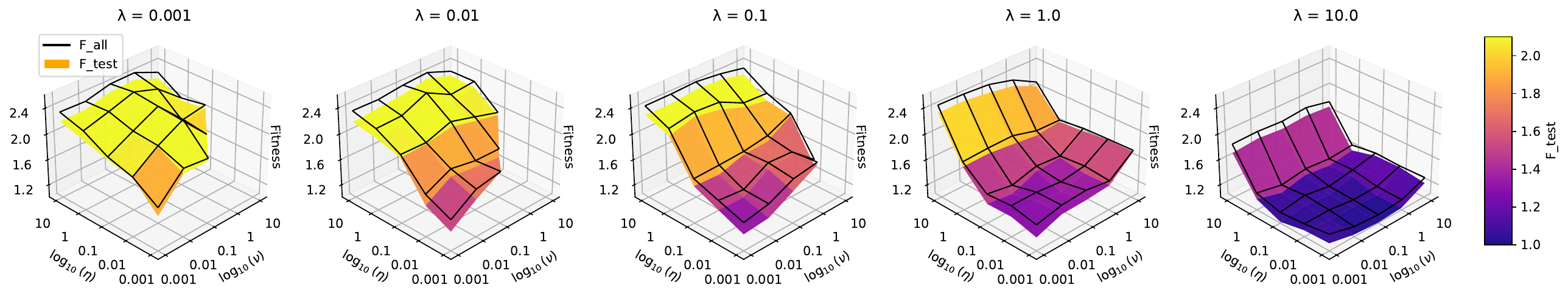}
        \caption{\textcolor{black}{Diversity-DeepSAD using HT features.}}
        \label{fig:deepsad_ht_sensitivity}
    \end{subfigure}

    \caption{\textcolor{black}{FUll hyperparameter sensitivity analysis of Diversity-DeepSAD over $(\nu, \eta, \lambda)$, showing response surfaces for the two SP methods.}}
    \label{fig:deepsad_sensitivity_overall}
\end{figure}

\begin{figure}[H]
    \centering

    \begin{subfigure}{\linewidth}
        \centering
        \includegraphics[width=\linewidth]{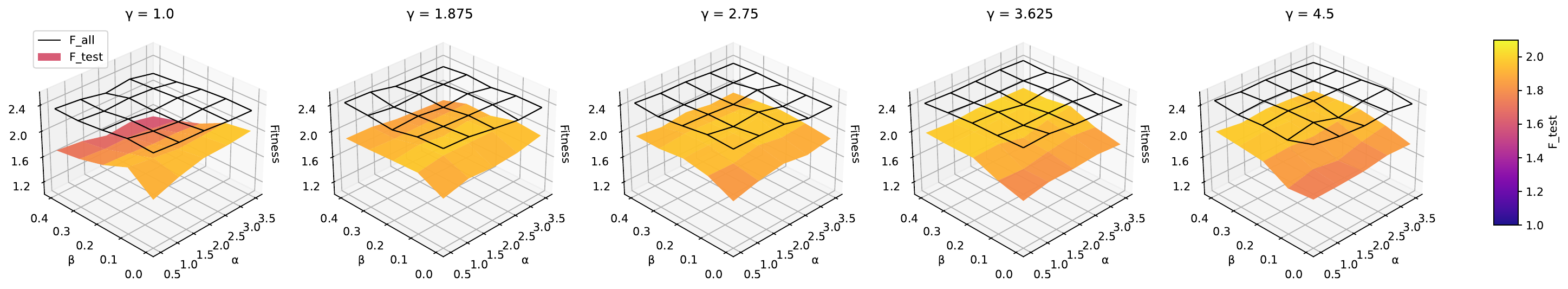}
        \caption{\textcolor{black}{DTC-VAE using FFT features.}}
        \label{fig:vae_fft_sensitivity}
    \end{subfigure}

    \vspace{8pt}

    \begin{subfigure}{\linewidth}
        \centering
        \includegraphics[width=\linewidth]{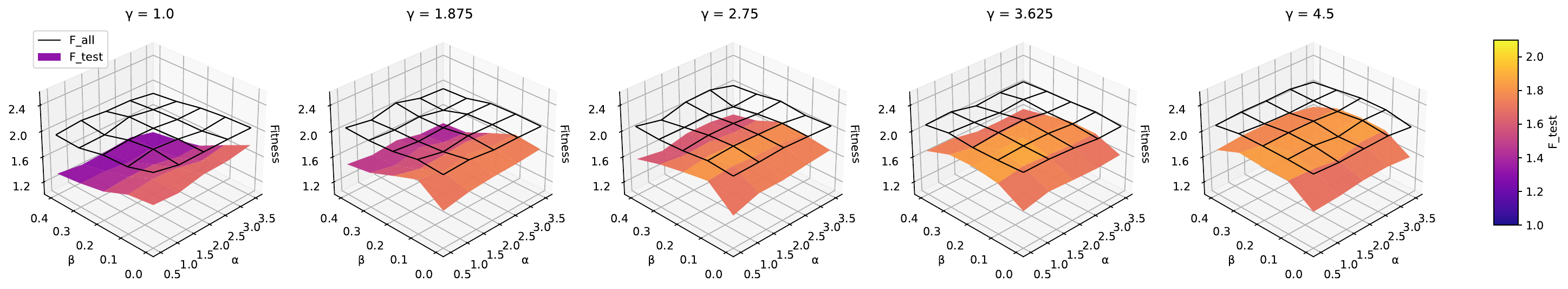}
        \caption{\textcolor{black}{DTC-VAE using HT features.}}
        \label{fig:vae_ht_sensitivity}
    \end{subfigure}

    \caption{\textcolor{black}{Full hyperparameter sensitivity analysis of DTC-VAE over $(\alpha, \beta, \gamma)$, showing response surfaces for the two SP methods.}}
    \label{fig:vae_sensitivity_overall}
\end{figure}

\noindent \textcolor{black}{All plots for the sensitivity analysis described in \autoref{subsec:sensitivity_method} are included below in \autoref{fig:deepsad_sensitivity_overall} and \autoref{fig:vae_sensitivity_overall}.}

\section{Code Availability}\label{app: code}
\noindent \textcolor{black}{The source code used for all experiments in this study is available on GitHub at the following repository:}
\begin{center}
\textcolor{black}{\hyperref[https://github.com/mortezamkh1992/Diversity-DeepSAD-vs-DTC-VAE]{https://github.com/mortezamkh1992/Diversity-DeepSAD-vs-DTC-VAE}}
\end{center}

\end{document}